\documentclass[lettersize,journal]{IEEEtran}
\usepackage{amsmath,amsfonts}
\usepackage{algorithmic}
\usepackage{array}
\usepackage[caption=false,font=normalsize,labelfont=sf,textfont=sf]{subfig}
\usepackage{textcomp}
\usepackage{stfloats}
\usepackage{url}
\usepackage{verbatim}
\usepackage{graphicx}
\usepackage{amssymb}
\usepackage[switch]{lineno}

\def\BibTeX{{\rm B\kern-.05em{\sc i\kern-.025em b}\kern-.08em
  T\kern-.1667em\lower.7ex\hbox{E}\kern-.125emX}}
\usepackage{balance}
\usepackage{multirow}

\usepackage[colorlinks,
      linkcolor=blue,
      anchorcolor=blue,
      citecolor=blue]{hyperref}

\begin{document}

\title{Transformer-based RGB-T Tracking with Channel and Spatial Feature Fusion}

\author{Yunfeng Li, Bo Wang, Ye Li
\thanks{Corresponding author: Bo Wang. \par The authors are with the National Key Laboratory of Autonomous Marine Vehicle Technology, Harbin Engineering University, Harbin 150001, China (email: liyunfeng@hrbeu.edu.cn, cv\_heu@163.com, liye@hrbeu.edu.cn).} }

\maketitle

\begin{abstract}
\textcolor{black}{The main problem in RGB-T tracking is the correct and optimal merging of the cross-modal features of visible and thermal images. Some previous methods either do not fully exploit the potential of RGB and TIR information for channel and spatial feature fusion or lack a direct interaction between the 
template and the search area, which limits the model’s ability to fully utilize the original semantic 
information of both modalities. To address these limitations, we investigate how to achieve a direct 
fusion of cross-modal channels and spatial features in RGB-T tracking and propose CSTNet. It uses 
the Vision Transformer (ViT) as the backbone and adds a Joint Spatial and Channel Fusion Module (JSCFM) and Spatial Fusion Module (SFM) integrated between the transformer blocks to facilitate 
cross-modal feature interaction.} The \textcolor{black}{JSCFM} module \textcolor{black}{achieves joint modeling of channel and multi-level spatial features}. The SFM module \textcolor{black}{includes} a cross-attention-like architecture \textcolor{black}{for cross modeling and joint learning of RGB and TIR features}. Comprehensive experiments show that CSTNet achieves state-of-the-art performance. To \textcolor{black}{enhance} practicality, we retrain the model without \textcolor{black}{JSCFM} and SFM modules and use CSNet as the pretraining weight, and propose CSTNet-small, which achieves 50\% speedup with an average \textcolor{black}{decrease of} 1-2\% in SR and PR performance. CSTNet and CSTNet-small achieve real-time speeds of 21 \textit{fps} and 33 \textit{fps} on the Nvidia Jetson Xavier, meeting actual deployment requirements. Code is available at \url{https://github.com/LiYunfengLYF/CSTNet}.

\end{abstract}


\section{Introduction}

 RGB-Thermal (RGB-T) tracking combines the advantages of the visible modality, which contains rich color and texture information, and the infrared modality, which is less affected by the environment, to achieve stable target tracking in low-light and adverse weather conditions. RGB-T Tracking aims to initialize the tracker through visual and infrared appearance templates of the target, and predict the target position and scale in subsequent bimodal frames. It has a wide range of potential applications in traffic monitoring \cite{backgroundvideo}, robot perception \cite{backgroundrobot}\textcolor{black}{,} and autonomous driving \cite{backgroundautodriving}. In these applications, improving the performance and efficiency of RGB-T trackers is crucial to improve target monitoring.

\begin{figure}
	\centering
    \includegraphics[width=8.0cm]{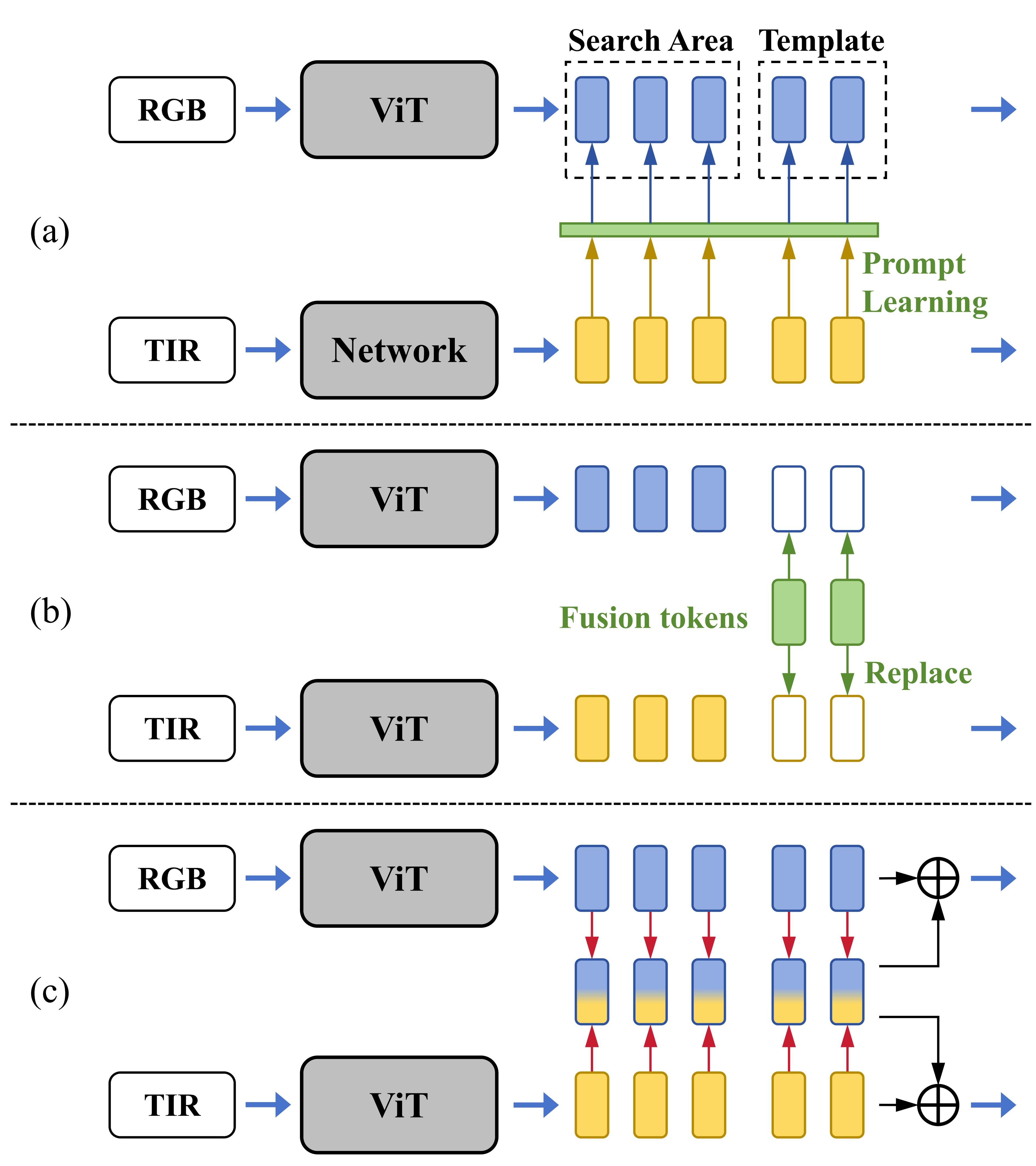}
	\caption{Comparison between our method with other Transformer-based RGB-T trackers. (a) and (b) represents the indirect fusion method currently used by the tracker. (a) TIR features are fused into RGB features by extracting prompts or adapter feature, such as ViPT \cite{vipt} and BAT \cite{bat}. (b) RGB and TIR features interact through the intermediate tokens, such as TBSI \cite{tbsi}. (c) Our model explores a method for direct interaction of cross-modal features, which is significantly different from the indirect feature interaction trackers represented by (a) and (b).}
	\label{fig:fig1}
\end{figure}

As an \textcolor{black}{RGB-T} vision task, the \textcolor{black}{high-quality} interaction of cross-modal features is crucial.
The main paradigms of RGB-T Transformer trackers currently include module insertion (e.g. TBSI \cite{tbsi}), prompt learning (e.g. ViPT \cite{vipt}) and insertion of the adapter layer (e.g. BAT \cite{bat}). 
Although these methods greatly improve the modeling of RGB and thermal infrared (TIR) features, as visualized in Figure \ref{fig:fig1} (a-b), all existing methods depend on intermediary modules for cross-modal interaction. The use of intermediaries introduces two critical bottlenecks: more layers for inference or large-scale downsampling of key information in features, which may result in distortion or loss of key features.
\color{black}

To alleviate this issue, we explore a direct feature interaction method for RGB-T tracking to avoid key feature loss caused by the use of intermediaries. Specifically, our goal is to maintain attention to the original cross-modal features during interaction by modeling spatial-channel characteristics of both RGB and TIR features, while achieving a smooth transition of the space and channel dimensions. 
Although prior work \cite{amnet} has investigated the impact of RGB-T image misalignment, we adopt the widely used assumption of spatial registration between modalities in this study.

Channel and spatial feature fusion is a conventional method for direct RGB-T feature interaction. MDNet-based RGB-T \cite{dafnet}\cite{cbpnet} trackers explore cross-modal feature interaction modules with spatial or channel attention as the core. However, these modules are more focused on the naive application of channel and spatial attention in RGB-T vision. Siamese-based RGB-T trackers \cite{isj_1}\cite{isj_2}\cite{isj_3}\cite{isj_4} further explore channel and spatial fusion methods, which achieve more refined channel or spatial attention and cross-attention-based feature modeling. However, these methods only focus on refining the modeling of spatial or channel features, lacking comprehensive and joint processing of both. In Transformer-based RGB-T trackers, such methods have not received widespread attention. Overall, current channel and spatial feature fusion methods have not fully utilized RGB and TIR features.

Therefore, this paper aims to combine \textcolor{black}{well-designed and comprehensive} channel and \textcolor{black}{spatial} feature fusion methods with Vision Transformer (ViT) to achieve direct interaction of \textcolor{black}{RGB and TIR features in transformer trackers. Although ViT provides a direct interaction paradigm for RGB and TIR features based on token concatenation and Multi-Head Self-Attention (MHSA), its computational complexity increases exponentially with the concatenation of RGB and TIR tokens. Therefore, inserting channels and spatial fusion modules is a compromise between the efficiency and performance of the feature direct interaction paradigm.}

\begin{figure}[t]
  \begin{minipage}[t]{0.5\linewidth}
    \centering
    \includegraphics[width=4.8cm]{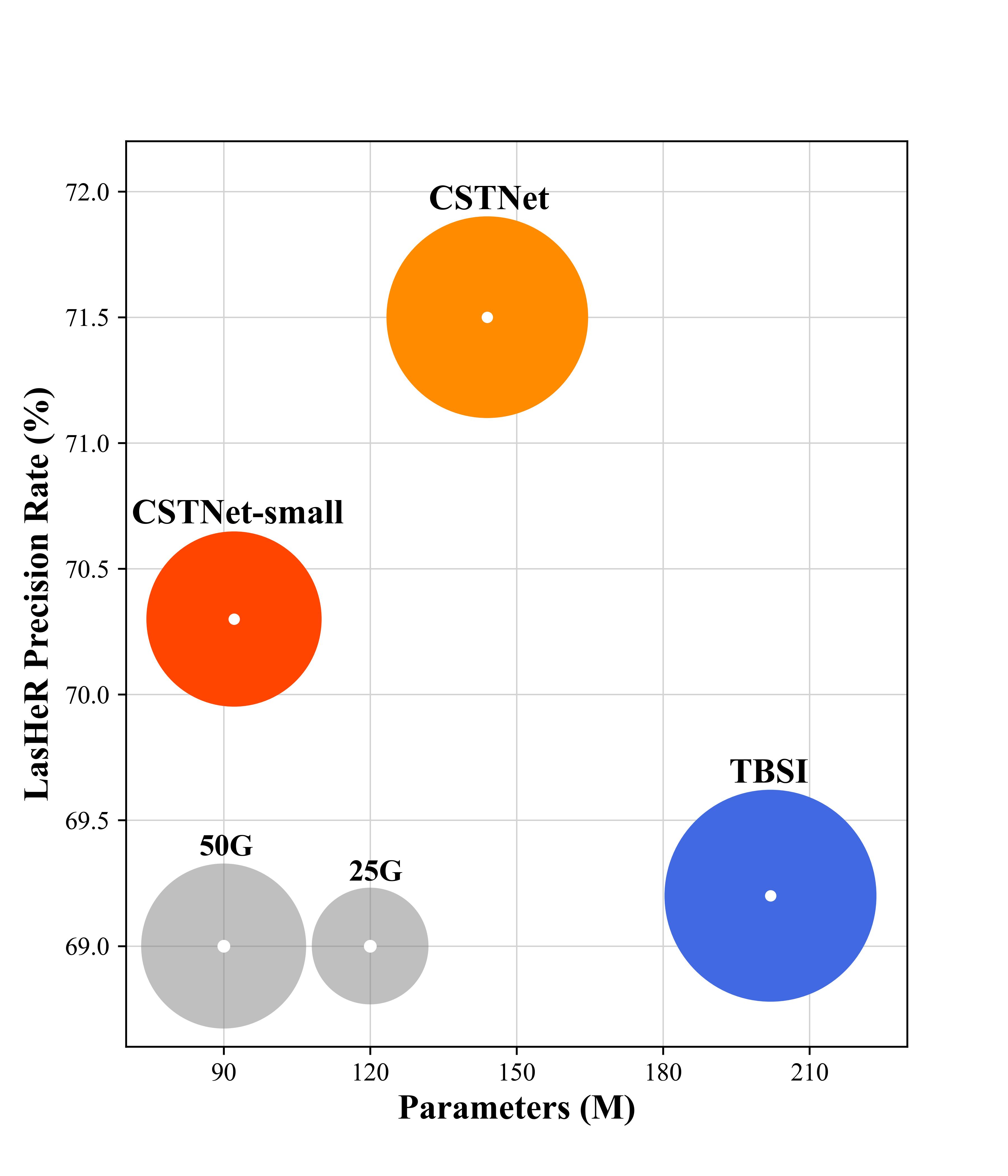}
  \end{minipage}%
  \begin{minipage}[t]{0.5\linewidth}
    \centering
    \includegraphics[width=4.8cm]{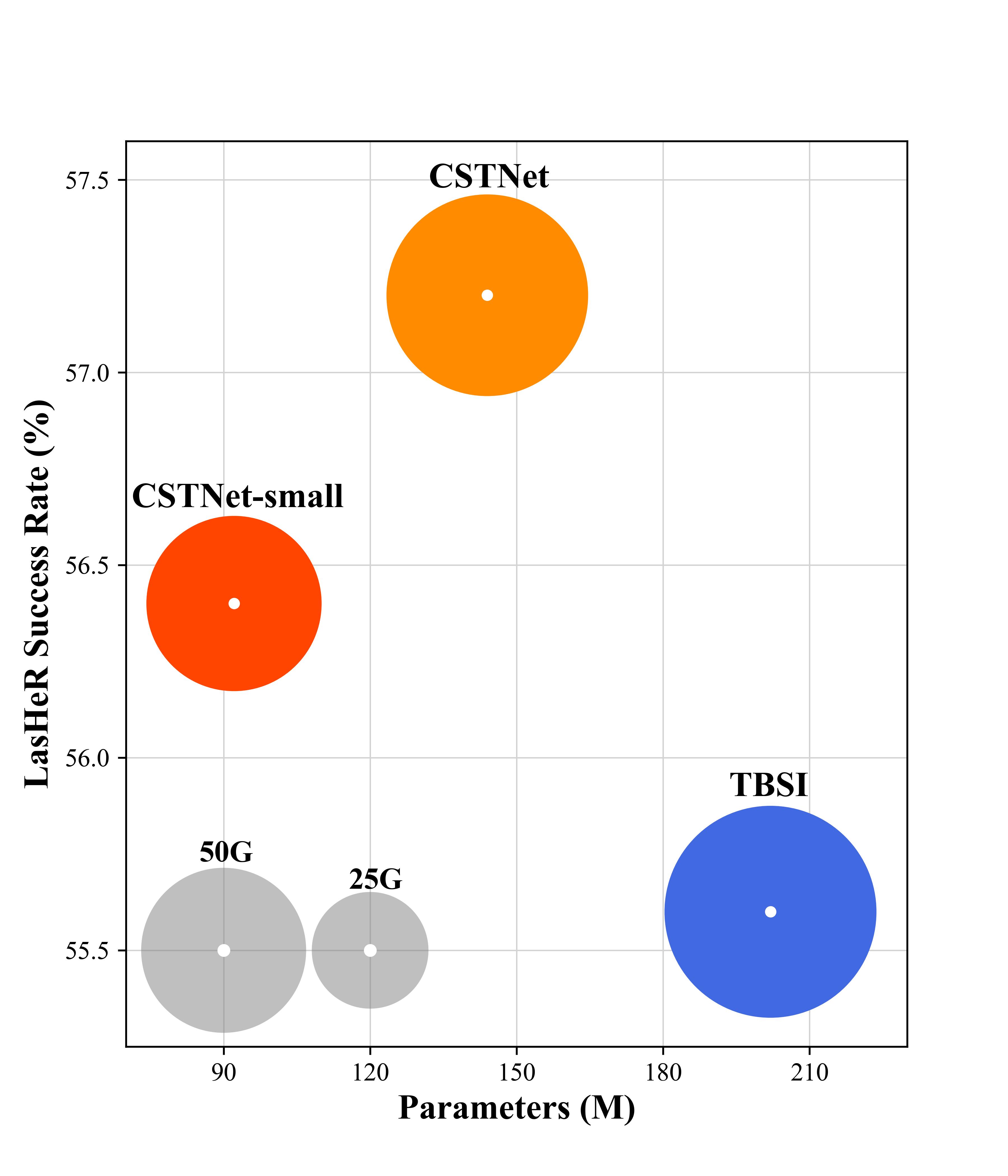}
  \end{minipage}
  \caption{Comparisons with our baseline TBSI \cite{tbsi} in terms of precision rate, success rate, parameters and \textcolor{black}{FLOPs} on \textcolor{black}{the} LasHeR \cite{lasher} benchmark. Our CSTNet and CSTNet-small reduce 28.7\% and 54.4\% in parameters, 9.5\% and 31.6\% in \textcolor{black}{FLOPs}, respectively. }
  \label{fig:figSRPR}
\end{figure}

In this work, we propose the \textcolor{black}{Channel and Spatial Transformer Network (CSTNet) for RGB-T tracking,} which leverages the advantages of directly integrating RGB and TIR features in both spatial and channel dimensions within ViT. \textcolor{black}{CSTNet contains} a \textcolor{black}{Joint Spatial and Channel Fusion Module (JSCFM)} and a Spatial Fusion Module (SFM). These modules are cascaded into a block and inserted into the ViT backbone. The \textcolor{black}{JSCFM} module \textcolor{black}{achieves direct interaction with RGB-T features by jointly modeling the channel and multi-scale spatial features at corresponding pixel locations.} The SFM module \textcolor{black}{achieves direct feature interaction through cross modeling, feature reuse, and joint learning of RGB-T features.} \textcolor{black}{We train the model on the LasHeR \cite{lasher} training set and evaluate it on the LasHeR \cite{lasher} test set, GTOT \cite{gtot}, RGBT210 \cite{rgbt210}, RGBT234 \cite{rgbt234}, VTUAV \cite{vtuav}, and UniRTL \cite{unirtl}.}

\textcolor{black}{To the best of our knowledge, our work is the first to introduce direct interactions in ViT-based RGB-T trackers. We introduce JSCFM and SFM modules to continuously model RGB-T features through joint-modeling, cross-modeling, and feature reuse, allowing the model to directly focus on the template and search area tokens themselves. Our method preserves key information in templates and search area tokens by avoiding the spatial or channel dimension reduction caused by intermediaries.}

To enhance the model practicality, we introduce a scaled-down version of CSTNet, denoted as CSTNet-small. \textcolor{black}{It removes} the \textcolor{black}{JSCFM} and SFM modules, utilizing the CSTNet model as the pre-training weights. The streamlined CSTNet-small, featuring reduced parameters and enhanced speed, as shown in Figure \ref{fig:figSRPR}, is poised to facilitate the implementation of our research in diverse applications, including intelligent vehicles such as uncrewed aerial vehicles (UAVs), uncrewed surface vehicles (USVs), underwater vehicles (UVs), and more.

Our main contributions are summarized as follows:

(1) We present a \textcolor{black}{Joint Spatial and Channel Fusion Module (JSCFM)} that concatenates and linearly fuses RGB and TIR features. \textcolor{black}{It} performs channel enhancement and multi-level spatial feature modeling in parallel, sums the output features, and globally integrates the hybrid features with the original features.

(2) We present a \textcolor{black}{S}patial \textcolor{black}{F}usion \textcolor{black}{M}odule (SFM) that aggregates local spatial features, models the spatial relationships of multi-modal features using a cross-attention mechanism, and integrates the spatial and channel information of multi-modal features through a convolutional feedforward network.

(3) We develop RGB-T trackers named CSTNet and CSTNet-small. The former effectively combines channel and spatial feature fusion with\textcolor{black}{in} the \textcolor{black}{ViT} model. The latter further reduces model parameters and improves speed, enhancing the practicality of the model. Comprehensive experiments demonstrate that our method\textcolor{black}{s} achieve state-of-the-art performance.

\section{Related Work}

\subsection{RGB-T tracker based on direct Feature Interaction}
Some RGB-T trackers propose channel fusion and spatial fusion methods for cross-modal features. DAFNet \cite{dafnet} proposes a channel-attention-based recursive fusion chain to achieve cross-modal and multi-level feature fusion in MDNet. SiamCDA \cite{siamcda} proposes a complementary\textcolor{black}{-}aware feature fusion module. It concatenates RGB and TIR features, and then enhances cross-modal features through a joint channel attention module. Finally, it achieves feature fusion through cross-modal residual connection\textcolor{black}{s} and feature concatenation. SiamTDR \cite{siamtdr} concatenates RGB and TIR features and then calculates the weight of channel attention \textcolor{black}{using} dynamic convolution to obtain fused feature. The fusion module of SiamMALL \cite{siammlaa} is similar to the above methods, except that it replaces the feature concatenation of the last layer with feature addition. MACFT \cite{MACFT} proposes a cross-attention fusion module and a mix-attention fusion module for spatial feature fusion in Transformer. It only performs cross-attention operations on the multimodal features output by its backbone to obtain fused features. \textcolor{black}{MCTrack \cite{mctrack} proposes an adaptive enhancement channel fusion module for bidirectional flow to achieve cross-modal channel enhancement of templates and search area\textcolor{black}{s}. STTrack \cite{isj_st1} proposes a state-based channel fusion module. APTrack \cite{isj_prompt1} explores a spatial attention method for cross-modal token modeling. }

\textcolor{black}{The above methods usually only integrate RGB-T features in the spatial or channel dimension, lacking the modeling of channel-to-spatial correlations. In contrast, our JSCFM and SFM modules, through different forms of spatial and channel joint integration, achieve deep mining of key information of RGB-T features from the channel and spatial dimensions. In addition, the cascading use of the two modules demonstrates the comprehensiveness of our method in integrating cross-modal spatial and channel features.}

\subsection{RGB-T tracker based on Indirect Feature Interaction}

BAT \cite{bat} proposes a universal bi-directional adapter to enable interaction between \textcolor{black}{the} RGB and TIR features in transformer blocks. TBSI \cite{tbsi} utilizes the fused template as a bridge for information interaction between RGB and TIR modalities in the search area, based on the attention mechanism. CAT++ \cite{catpp} introduces five attribute-based branches to address different challenges and fuses features from these branches through an aggregation interaction module for multimodal tracking. TATrack \cite{tatrack} employs an online template to model temporal information and \textcolor{black}{facilitate} cross-modal feature interaction between the initial and online templates. 

Although these methods achieve \textcolor{black}{RGB-T} feature fusion through intermediate features, prior knowledge, and temporal information, they do not explore the effectiveness of direct interaction between RGB and TIR features. This limitation leads to the model's inability to leverage the original semantic \textcolor{black}{interactions} between RGB and TIR features. In contrast, our method delves into the potential for direct channel and spatial fusion of cross-modal features within the transformer network.

\subsection{Feature fusion for other RGB-T visual tasks}
The channel and spatial feature fusion module has been widely applied to non-sequential RGB-T tasks. CMX \cite{cmx} employs cross-modal channel and spatial attention to complement and refine multimodal features. It introduces a feature fusion module that integrates a cross-attention mechanism with a convolutional feedforward layer for spatial modeling. MMSFormer \cite{mmsformer} presents a multimodal feature fusion module that enhances feature representation through multi-level fusion and channel enhancement, fusing multimodal features from various backbone stages to create a robust mixed-modality representation. \textcolor{black}{HGT \cite{isj_tmm1} explores an RGB-T feature interaction module that combines graph networks with Transformers. PLASSO-ADSPF \cite{isj_kbs1} explores \textcolor{black}{an} RGB-T feature quality-aware fusion method that incorporates both deep and handcrafted features.}

Compared to these methods, our method differs in terms of task formulation. In comparison to the cross-attention module in CMX \cite{cmx}, our cross-attention-like module enhances cross-modal feature interaction through the following key improvements: aggregating local spatial features, sharing integration layers, maintaining consistency in cross-modal feature channels during fusion, and decoupling the learning of spatial and channel features for the target.

\section{Method}

\begin{figure*}
	\centering
    \includegraphics[width=18cm]{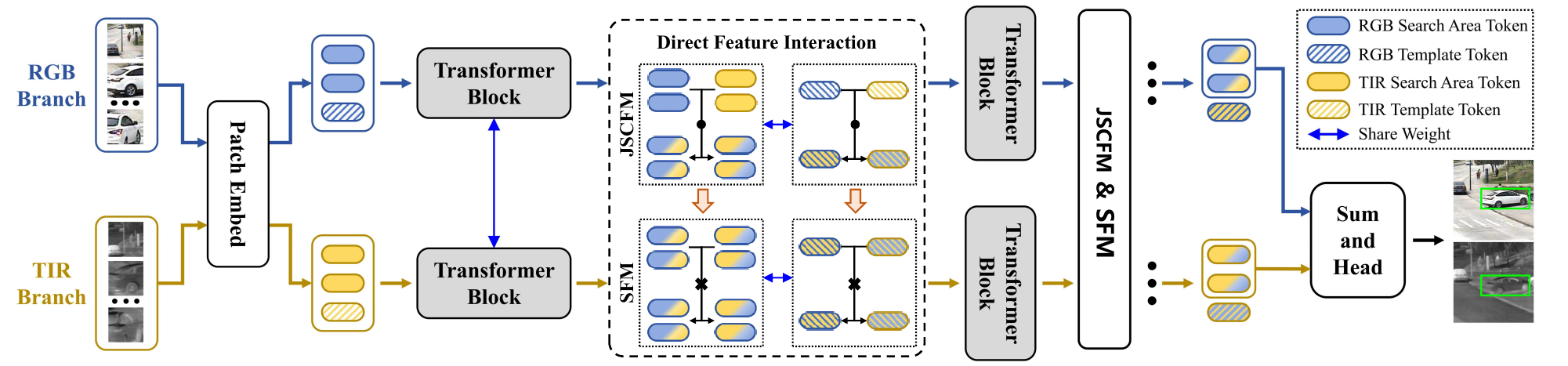}
	\caption{The overall framework of CSTNet. Our backbone is a bilateral ViT with shared weights. The RGB and TIR image patches are first embedded into tokens and fed into 12 Transformer Blocks. The proposed proposed \textcolor{black}{Joint Spatial and Channel Fusion Module (JSCFM)} and \textcolor{black}{Spatial Fusion Module (SFM)} perform cross-modal feature fusion on template features and search area features, respectively. They are inserted into 4-th, 7-th, and 10-th layers of the backbone. Finally, the RGB and TIR features of the search area are added, and the prediction head is used to predict the current state of the target.}
	\label{fig:fig2}
\end{figure*}

Our method is based on the assumption of spatial alignment between RGB \textcolor{black}{and TIR} images. The overall framework of CSTNet is illustrated in Figure \ref{fig:fig2}. We start by embedding the template images and search area images into tokens, which are then passed through \textcolor{black}{ViT} blocks. Our proposed \textcolor{black}{Joint Spatial and Channel Fusion Module (JSCFM)} and \textcolor{black}{Spatial Fusion Module (SFM)} are integrated between the ViT blocks to facilitate \textcolor{black}{RGB-T} feature interaction. Subsequently, the RGB and TIR features of the search area are added and input to the prediction head to predict the target state. Since \textcolor{black}{the JSCFM} and SFM modules process template and search area features identically, we \textcolor{black}{only display the feature processing of the search area} in Sections \ref{cfm} and \ref{sfm}. Additionally, CSTNet-small excludes the \textcolor{black}{JSCFM} and SFM modules from CSTNet, utilizing the CSTNet weights \textcolor{black}{as pre-training weights}.

\subsection{Backbone}
We \textcolor{black}{take} ViT \cite{ostrack} as our backbone, following recent RGB-T trackers \cite{tbsi}\cite{tatrack}. The backbone takes RGB and TIR template images as inputs, denoted as $z_{r}, z_{t} \in \mathbb{R}^{H_{z}\times W_{z}\times 3}$ and RGB and TIR search area images represented as $x_{r}, x_{t}\in \mathbb{R}^{H_{x}\times W_{x}\times 3}$. The $z_{r}, z_{t}, x_{r}, x_{t}$ are initially embedded as tokens, with positional embeddings added as follows:
\begin{equation}
\begin{split}
Z_{r}&=\text{PE}(z_{r}) + E_{z}, Z_{t}=\text{PE}(z_{t}) + E_{z} \\
X_{r}&=\text{PE}(x_{r}) + E_{x}, X_{t}=\text{PE}(x_{t}) + E_{x}
\end{split}
\tag*{(1)} 
\end{equation}
where PE is the Patch Embed function as \textcolor{black}{described} in \cite{ostrack}. $Z_{r}, Z_{t}\in \mathbb{R}^{N_{z}\times C}$ denotes the RGB and TIR template tokens. $N_{z}=H_{z}W_{z}/p^{2}$, where $p$ is the down-sampling stride of the patch embedding. $C$ is the \textcolor{black}{number of channels}. $X_{r}, X_{t}\in \mathbb{R}^{N_{x}\times C}$ denote the RGB and TIR \textcolor{black}{search} area tokens, where $N_{x}=H_{x}W_{x}/p^{2}$. $E_{z}$ and $E_{x}$ are the position \textcolor{black}{embeddings} with the same shape as $Z_{r}$ and $X_{r}$, respectively.
  
Since the feature extraction and modeling \textcolor{black}{are} the same for both modalities, we \textcolor{black}{will} only discuss the process in the RGB modality. The RGB tokens are concatenated as $H_{r}=[X_{r};Z_{r}] \in \mathbb{R}^{(N_{x}+N_{z})\times C}$, \textcolor{black}{and} then $H_{r}$ is fed into a series of \textcolor{black}{ViT} \cite{ostrack} blocks.

The proposed cross-modal feature fusion method is represented as $Z_{rt}=f(Z_{r}, Z_{t})$ and $X_{rt}=f(X_{r}, X_{t})$, where $Z_{rt}$ denotes the \textcolor{black}{fused} feature of the template and $X_{rt}$ denotes the fusion feature of the search area. After the fusion, the relationship modeling is represented as \textcolor{black}{follows}:
\begin{equation}
\small
\begin{split}
  A&= \text{Softmax}(\frac{QK^{T}}{\sqrt{C}})V=\text{Softmax}(\frac{[X_{qrt};Z_{qrt}][X_{krt};Z_{krt}]^{T}}{\sqrt{C}})V \\
   & =\text{Softmax}(\frac{[X_{qrt}X_{krt}^{T},X_{qrt}Z_{krt}^{T};Z_{qrt}X_{krt}^{T},Z_{qrt}Z_{krt}^{T}]}{\sqrt{C}})V
\end{split}
\tag*{(2)} 
\end{equation}
where \textcolor{black}{$Q$, $K$, $V$} are Query, Key, and Value matrices, respectively. 

From the above formulation, after the fusion, the template features and the search area features contain both RGB and TIR semantic information for relationship modeling. The joint modeling of these cross-modal features enhances the model's ability to utilize multi-level semantic information from both modalities.

\subsection{\textcolor{black}{Joint Spatial} and Channel Fusion Module}
\label{cfm}

\begin{figure}
	\centering
    \includegraphics[width=8.3cm]{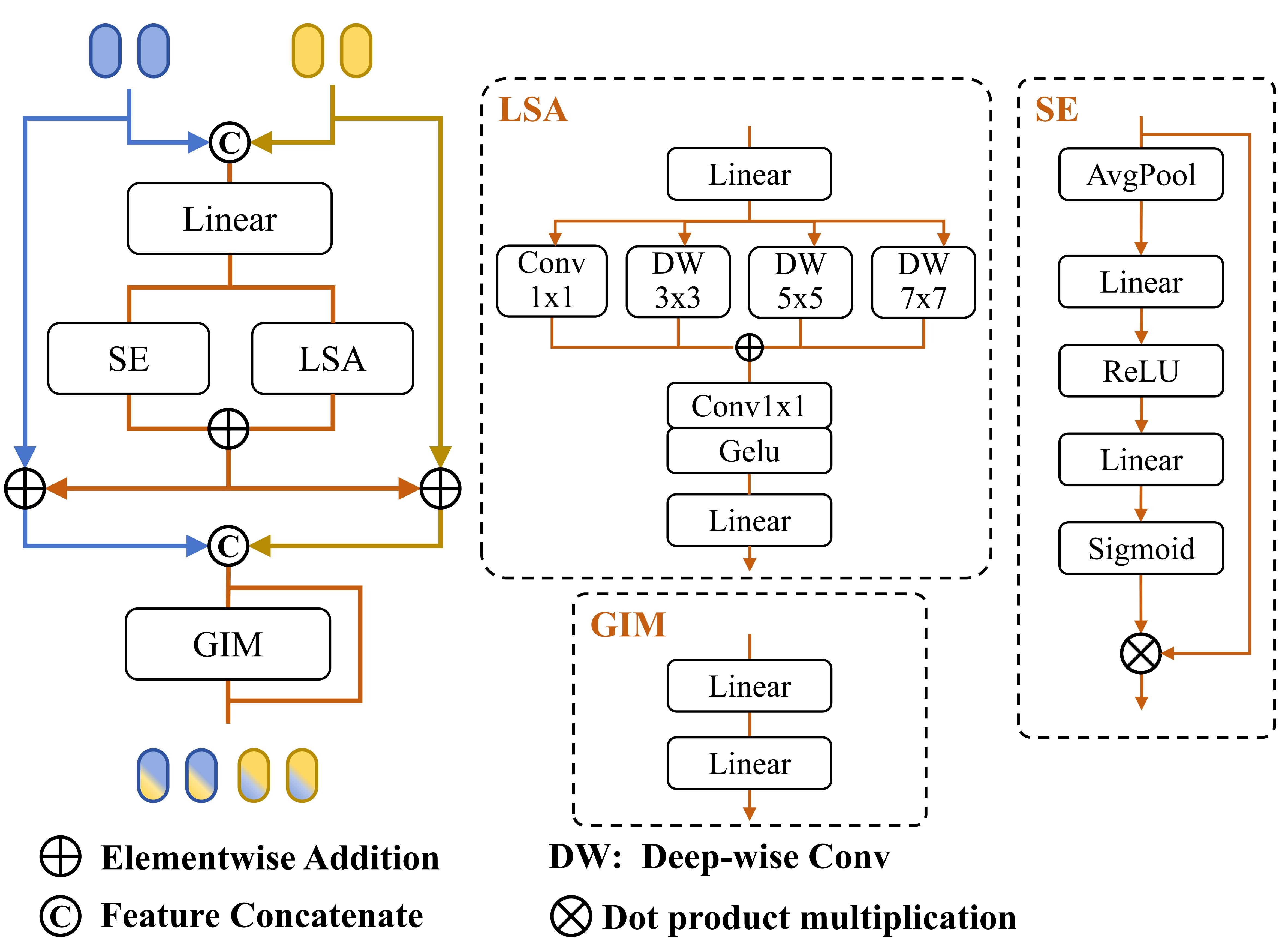}
	\caption{Illustration of our proposed \textcolor{black}{JSCFM} module. It encompasses \textcolor{black}{a Linear layer}, a\textcolor{black}{n Squeeze-and-Excitation (SE)} \cite{se} module, a \textcolor{black}{L}ocal \textcolor{black}{S}patial \textcolor{black}{A}ggregation (LSA) module, and a \textcolor{black}{G}lobal \textcolor{black}{I}ntegration \textcolor{black}{M}odule (GIM).}
	\label{fig: cfm}
\end{figure}

Our proposed \textcolor{black}{Joint Spatial} and Channel Fusion Module (\textcolor{black}{JSCFM}) \textcolor{black}{aims to jointly integrate the channel and spatial features of RGB and TIR modalities.} As illustrated in Figure \ref{fig: cfm}, the \textcolor{black}{JSCFM module} comprises \textcolor{black}{a linear layer}, an \textcolor{black}{Squeeze-and-Excitation (SE) module} for enhancing key information within channels, a Local Spatial Aggregation (LSA) module for collectively modeling multi-level spatial feature representations, and a Global Integration Module (GIM) for integrating the original and fused features.

We first linearly fuse multimodal features by concatenating the RGB and TIR features of the search area, \textcolor{black}{represented as:}
\begin{equation}
X_{c}=\text{Linear}([X_{r};X_{t}])\in \mathbb{R}^{N_{x}\times C}
\tag*{(3)} 
\end{equation}
where $\text{Linear}$ represents the linear layer. 

The \textcolor{black}{SE module} \cite{se} is employed to enhance the channels of $X_{c}$, \textcolor{black}{represented as:} 
\begin{equation}
\small
\begin{split}
X_{c}^{attn} = & \text{Sigmoid}(\text{Linear}({\text{ReLU}(\text{Linear(}{\text{AvgPool}(X_{c})})})) )\\
X_{c}^{\text{se}} = & X_{c}^{attn}\times X_{c}
\end{split}
\tag*{(4)} 
\end{equation}
where $\text{AvgPool}$ is adaptive average pooling. 

The \textcolor{black}{LSA module} is utilized to extract multi-level local spatial features and model key representations of mixed multi-level features through a non-linear layer. Specifically, a linear transformation is initially applied to the feature $X_{c}$, resulting in $X_{c}^{\text{fc}} = \text{Linear}(X_{c}) \in \mathbb{R}^{N_{x}\times C}$. Then we use convolutional blocks to extract multi-level spatial features, represented as:
\begin{equation}
X_{c}^{\text{msf}} = \text{Conv}_{1\times 1}(X_{c}^{\text{fc}})+ \sum_{k\in (3,5,7)}\text{DWConv}_{k\times k}(X_{c}^{\text{fc}})
\tag*{(5)} 
\end{equation}
where $\text{Conv}$ and $\text{DWConv}$ represent convolutional layer with batch-normalize (BN) and depth-wise separable convolutional layer with BN, respectively.

Another linear layer is used to perform a linear transformation on the fused features to integrate multi-level features and improve their consistency, represented as $X_{c}^{\text{lsa}} = \text{Linear}(\text{Gelu}(\text{Conv}_{1\times 1}(X_{c}^{\text{msf}}))) \in \mathbb{R}^{N_{x}\times C}$, where $\text{Gelu}$ represents the GELU \cite{gelu} \textcolor{black}{activation} function.

After the parallel SE and LSA modules, the features are added, represented as $X_{c}^{\text{add}}=X_{c}^{\text{se}}+X_{c}^{\text{lsa}} \in \mathbb{R}^{N_{x}\times C}$. In addition, residual connections are introduced to complement the original semantic information lost in \textcolor{black}{RGB-T} features, represented as $X_{r}^{\text{res}}=X_{r}+X_{c}^{\text{add}}$ and $X_{t}^{\text{res}}=X_{t}+X_{c}^{\text{add}}$.

The \textcolor{black}{GIM module} concatenates the RGB and TIR features and feeds \textcolor{black}{them} into a Multi-Layer Perceptron (MLP) with \textcolor{black}{a} residual connection. Then, the concatenated feature is split to obtain integrated RGB and TIR features. The process is described as follow:
\begin{equation}
  X_{r}^{\text{cfm}},X_{t}^{\text{cfm}}=\text{Split}([X_{r}^{\text{res}};X_{t}^{\text{res}}]+\text{MLP}([X_{r}^{\text{res}};X_{t}^{\text{res}}]))
\tag*{(6)} 
\end{equation}
where $\text{Split}$ maps the feature from $\mathbb{R}^{N_{x}\times 2C}$ to two $\mathbb{R}^{N_{x}\times C}$ features. $\text{MLP}$ consists of two linear layers. The $X_{r}^{\text{cfm}}$ and $X_{t}^{\text{cfm}}$ are the output features of the \textcolor{black}{JSCFM module}.

Due to the difference\textcolor{black}{s} in feature size and target location between the template and the search area, \textcolor{black}{the} GIM \textcolor{black}{module} is not shared between the template and the search \textcolor{black}{area} when processing features.

\subsection{Spatial \textcolor{black}{F}usion \textcolor{black}{M}odule}
\label{sfm}

\begin{figure}
	\centering
    \includegraphics[width=7.8cm]{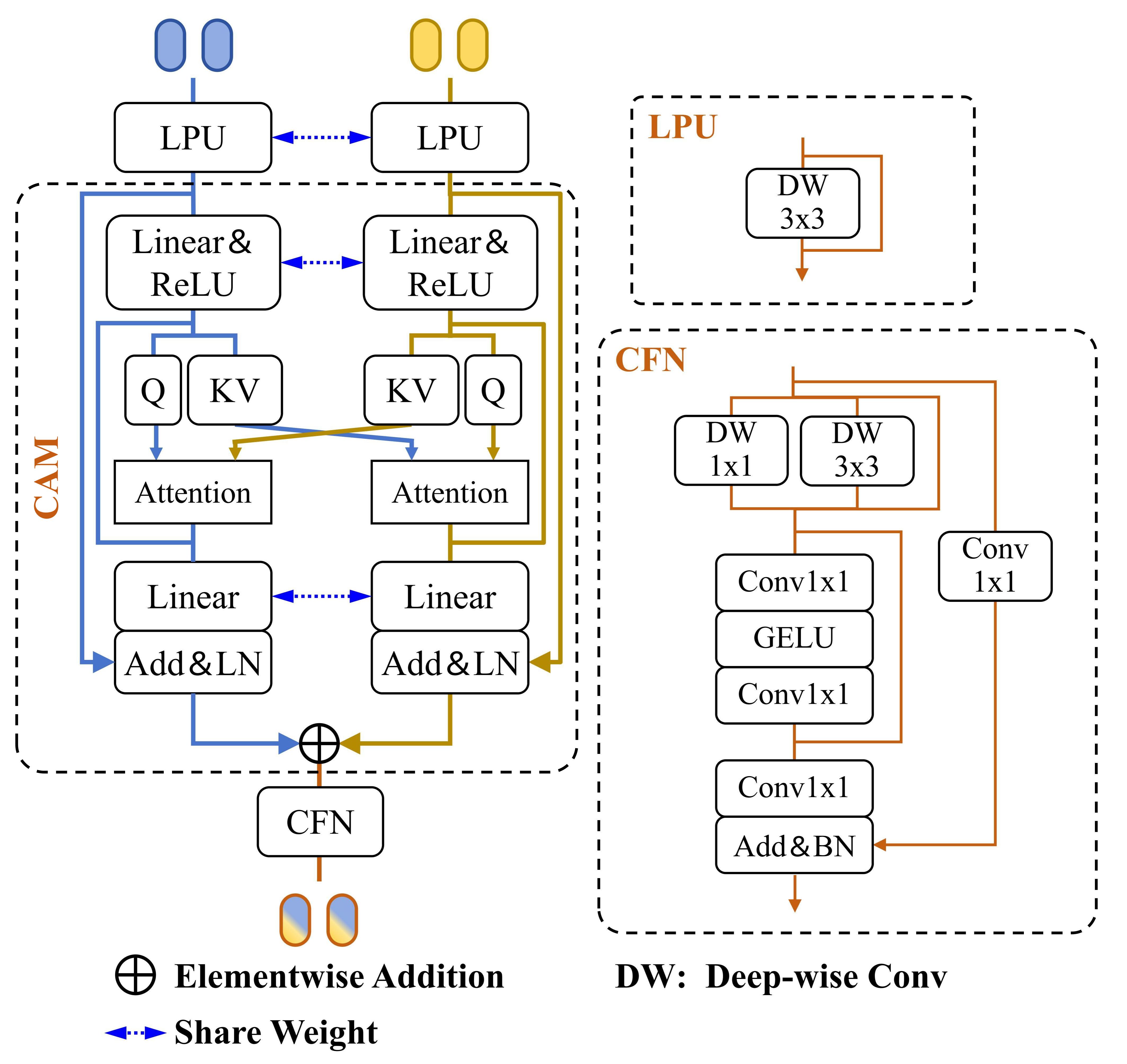}
	\caption{Illustration of our proposed SFM module. It encompasses a \textcolor{black}{L}ocal \textcolor{black}{P}erception \textcolor{black}{U}nit (LPU), a \textcolor{black}{C}ross-\textcolor{black}{A}ttention \textcolor{black}{M}odule \textcolor{black}{(CAM)}, and a \textcolor{black}{C}onvolutional \textcolor{black}{F}eedforward \textcolor{black}{N}etwork (CFN).}
	\label{fig:sfm}
\end{figure}

Our proposed \textcolor{black}{Spatial Fusion Module (SFM)} module aims to model spatial relationships in cross-modal features and integrate the channel and spatial features of the RGB and TIR modalities through a convolutional feedforward network. As shown in Figure \ref{fig:sfm}, the SFM \textcolor{black}{module} contains a \textcolor{black}{L}ocal \textcolor{black}{P}erception \textcolor{black}{U}nit (LPU), a {\textcolor{black}C}ross-\textcolor{black}{A}ttention \textcolor{black}{M}odule \textcolor{black}{(CAM)}, and a {\textcolor{black}C}onvolutional \textcolor{black}{F}eedforward \textcolor{black}{N}etwork (CFN).

\textcolor{black}{The LPU module is used} to aggregate local spatial information, represented as:
\begin{equation}
\begin{split}
  X_{r}^{\text{lpu}}=X_{r}^{\text{cfm}}+\text{DWConv}_{3\times 3}(X_{r}^{\text{cfm}}) \\
  X_{t}^{\text{lpu}}=X_{t}^{\text{cfm}}+\text{DWConv}_{3\times 3}(X_{t}^{\text{cfm}}) 
\end{split}
\tag*{(7)} 
\end{equation}

\textcolor{black}{The CAM module} is used to achieve spatial modeling of cross-modal features. Specifically, a non-linear full \textcolor{black}{connected} layer is used and represented as $X_{r}^{\text{res}}, X_{r}^{\text{attn}}= \text{Split}(\text{Relu}(\text{Linear}^{\text{s}}(X_{r}^{\text{lpu}})))$, where $\text{Linear}^{\text{r}}$ is a linear layer shared by \textcolor{black}{the} RGB and TIR branches. After that, a linear layer is used to obtain Query: $X_{r}^{q}=\text{Linear}(X_{r}^{\text{attn}})$, another linear layer is used to obtain Key: $X_{r}^{k}$ and Value: $X_{r}^{v}$, represented as $X_{r}^{k}, X_{r}^{v}= \text{Split}(\text{Linear}(X_{r}^{\text{attn}})$. The TIR $X_{t}^{q}, X_{t}^{k}, X_{t}^{v}$ are obtained in the same way. The cross-attention process of RGB and TIR features is described as follows:
\begin{equation}
\begin{split}
  X_{r}^{cross}= \text{Attention}(Q:X_{r}^{q}, K:X_{t}^{k}, V:X_{t}^{v}) \\
  X_{t}^{cross}= \text{Attention}(Q:X_{t}^{q}, K:X_{r}^{k}, V:X_{r}^{v}) \\
\end{split}
\tag*{(8)} 
\end{equation}

Then we use residual connections and a shared linear layer to adjust the features and sum \textcolor{black}{them}. 
The reason for using \textcolor{black}{sumsummation} instead of concatenation \textcolor{black}{is} to ensure that the features before and after the cross-attention interaction reflect the relationship in the spatial features, and \textcolor{black}{summation} ensures that there is no change in the feature channel structure. 
The shared layer improves the learning of unified features in \textcolor{black}{the} cross-modal features of the target. 
 The process is described as:
\begin{equation}
\begin{split}
  X_{r}^{\text{add}}&= \text{LN}_{r}(X_{r}^{\text{cfm}}+\text{Linear}^{s}(X_{r}^{cross}+X_{r}^{res})) \\
  X_{t}^{\text{add}} &= \text{LN}_{t}(X_{t}^{\text{cfm}}+\text{Linear}^{s}(X_{t}^{cross}+X_{t}^{res}))
\end{split}
\tag*{(9)} 
\end{equation}
where $\text{LN}_{r}$ and $\text{LN}_{r}$ are different layer-normalize layers.

For the same reasons \textcolor{black}{mentioned} above, we add the features $X_{r}^{\text{add}}$ and $X_{t}^{\text{add}}$ to achieve feature point-level interaction between the multimodal features.
\begin{equation}
X_{rt}^{\text{add}} =X_{r}^{\text{add}}+X_{t}^{\text{add}}
\tag*{(10)} 
\end{equation}

\textcolor{black}{The CFN module is introduced} due to the differences between the RGB and TIR features, the directly added feature requires further spatial and channel integration. \textcolor{black}{It} improves the spatial and channel representation of key features by learning \textcolor{black}{the} spatial and channel decoupling of cross-modal features. We first perform spatial integration on the merged feature, \textcolor{black}{represented as follows}:
\begin{equation}
  X_{rt}^{local} = X_{rt}^{\text{add}}+ \sum_{k\in (1,3)}\text{DWConv}_{k\times k}(X_{rt}^{\text{add}})
\tag*{(11)} 
\end{equation}

Then, we nonlinearly integrate the channel feature\textcolor{black}{s} of the merged feature, represented as \textcolor{black}{follows}:
\begin{equation}
  X_{rt}^{act} = X_{rt}^{local}+\text{Conv}_{1\times 1}^{\text{down}}(\text{Gelu}(\text{Conv}_{1\times 1}^{\text{up}}(X_{rt}^{local})))
\tag*{(12)} 
\end{equation}
where $\text{Conv}_{1\times 1}^{\text{up}}$ maps the feature from $\mathbb{R}^{C}$ to $\mathbb{R}^{2\times C}$\textcolor{black}{,} and $\text{Conv}_{1\times 1}^{\text{down}}$ maps the feature from $\mathbb{R}^{2\times C}$ to $\mathbb{R}^{C}$.

In addition, we use a 1$\times$1 convolutional block and a skip connection with \textcolor{black}{another} 1$\times$1 convolutional block to adjust the features. The process is described as follows:
\begin{equation}
  X_{rt}^{adj} = \text{BN}(\text{Conv}_{1\times 1}^{\text{adj}}(X_{rt}^{act})+\text{Conv}_{1\times 1}^{\text{res}}(X_{rt}^{act}))
\tag*{(13)} 
\end{equation}
where $\text{Conv}_{1\times 1}^{\text{adj}}$ and $\text{Conv}_{1\times 1}^{\text{res}}$ both map the feature\textcolor{black}{s} from $\mathbb{R}^{C}$ to $\mathbb{R}^{C}$.

Finally, we add $X_{rt}^{\text{sfm}}$ to the original RGB feature $X_{r}$ and the original TIR feature $X_{t}$ to obtain the fused features as \textcolor{black}{follows}:
\begin{equation}
  X_{r}^{\text{sfm}} = X_{rt}^{adj}+X_{r},\ X_{t}^{\text{sfm}} = X_{rt}^{adj}+X_{t}
\tag*{(14)} 
\end{equation}
where $X_{r}^{\text{sfm}}$ and $X_{t}^{\text{sfm}}$ are the output features of SFM \textcolor{black}{module}. 

\subsection{Head and Loss}

\textcolor{black}{The search area RGB-T features output by the backbone} are represented as $X_{r}^{\text{output}}$ and $X_{t}^{\text{output}}$. We add them \textcolor{black}{together} to obtain the mixed feature, described as  follows:
\begin{equation}
  X_{rt}^{\text{output}}=X_{r}^{\text{output}}+X_{t}^{\text{output}}
\tag*{(15)} 
\end{equation}
where $X_{rt}^{\text{output}}$ is the input of our prediction head.

Following the design of \cite{ostrack} and \cite{tbsi}, we adopt a fully convolutional center-head \cite{centerhead} to predict the state of the target. We use the \textcolor{black}{weighted} focal loss \cite{weightfocalloss} to train the classification branch, and \textcolor{black}{we use }the l1 loss and the wise-IoU loss \cite{wiou} to train the box regression branch. The total training loss of CSTNet is: 
\begin{equation}
  L_{total}=L_{cls}+\lambda_{iou}L_{iou}+\lambda_{l_{1}}L_{1}
\tag*{(16)} 
\end{equation}
where $\lambda_{iou}=2$ and $\lambda_{l_{1}}=5$ are hyper-parameters. We set the values of these two hyperparameters by following the current values used in most trackers (such as \cite{vipt}, \cite{tbsi}, \cite{bat}).
\section{Experiments}

\begin{table*}[t]
\caption{Comparison with state-of-the-art trackers on VTUAV \cite{vtuav}, RGBT210 \cite{rgbt210}, RGBT234 \cite{rgbt234}, and LasHeR \cite{lasher} testing set. Higher value Indicate better performance. The best two results
are shown in \color{red}{red} \color{black}{and} \color{blue}{blue} \color{black}{fonts}.}
\label{table rgbt234 and lasher reuslts}
\centering
\renewcommand{\arraystretch}{1.15}
\small %
\begin{tabular}{l|c|c|cc|cc|cc|ccc}
\hline
\multicolumn{1}{c|}{\multirow{2}{*}{Method}} & \multicolumn{1}{c|}{\multirow{2}{*}{Source}} & \multicolumn{1}{c|}{\multirow{2}{*}{Baseline}} & \multicolumn{2}{c|}{VTUAV \cite{vtuav}} & \multicolumn{2}{c|}{RGBT210 \cite{rgbt210}} & \multicolumn{2}{c|}{RGBT234 \cite{rgbt234}} & \multicolumn{3}{c}{LasHeR \cite{lasher}} \\ \cline{4-12} 
\multicolumn{1}{c|}{}                        & \multicolumn{1}{c|}{}                        & \multicolumn{1}{c|}{}                          & MPR          & MSR         & PR            & SR           & PR            & SR           & PR      & NPR     & SR     \\ \hline
TFNet \cite{tfnet}          & TCSVT2021       & MDNet             & -            & -           & 77.7          & 52.9         & 80.6          & 56           & -       & -       & -      \\
AGMINet \cite{agminet}      & TIM2022         & MDNet             & -            & -           & -             & -            & 84.0          & 59.2         & 48.8    & 42.9    & 34.3   \\
DMCNet \cite{dmcnet}        & TNNLS22         & MDNet             & -            & -           & 79.7          & 55.9         & 83.9          & 59.3         & 49.0    & 43.1    & 35.5   \\
APFNet \cite{apfnet}        & AAAI2022        & MDNet             & -            & -           & 79.9          & 54.9         & 82.7          & 57.9         & 50.0    & 43.9    & 36.2   \\
DMSTM \cite{dmstm}          & TIM2023         & MDNet             & -            & -           & -             & -            & 78.6          & 56.2         & 55.7    & 50.3    & 40.0   \\
X-Net \cite{xnet}           & Arxiv2023       & MDNet             & -            & -           & -             & -            & 85.2          & 62.2         & 50.8    & -       & 44.6   \\
QAT \cite{qat}              & ACM MM2023      & MDNet             & 80.1         & 66.7        & 86.8          & 61.9         & 88.4          & 64.4         & 64.2    & 59.6    & 50.1   \\
CAT++ \cite{catpp}          & TIP2024         & MDNet             & -            & -           & 82.2          & 56.1         & 84.0          & 59.2         & 50.9    & 44.4    & 35.6   \\ \hline
SiamCDA \cite{siamcda}      & TCSVT2021       & Siamese           & -            & -           & -             & -            & 76.0          & 56.9         & -       & -       & -      \\
DFAT \cite{DFAT}            & Inf Fusion23    & Siamese           & -            & -           & -             & -            & 75.8          & 55.2         & -       & -       & -      \\
SiamMLAA \cite{siammlaa}    & TMM2023         & Siamese           & -            & -           & -             & -            & 78.6          & 58.4         & -       & -       & -      \\ \hline
MFNet \cite{mfnet}          & IVC2022         & DCF               & -            & -           & -             & -            & 84.4          & 60.1         & 59.7    & 55.4    & 46.7   \\
HMFT \cite{vtuav}           & CVPR2022        & DCF               & 75.8         & 62.7        & 78.6          & 53.5         & 78.8          & 56.8         & -       & -       & -      \\
LSAR \cite{lsar}            & TCSVT2023       & DCF               & -            & -           & -             & -            & 78.4          & 55.9         & -       & -       & -      \\
CMD \cite{cmd}              & CVPR2023        & DCF               & -            & -           & -             & -            & 82.4          & 58.4         & 59.0    & 54.6    & 46.4   \\
MPT \cite{mpt}              & TIM2024         & DCF               & -            & -           & -             & -            & 76.9          & 55.0         & 35.5    & -       & 31.3   \\
AMNet \cite{amnet}          & TCSVT2024       & DCF               & -            & -           & -             & -            & 85.5          & 60.7         & 61.4    & 55.9    & 47.7   \\ \hline
ProTrack \cite{protrack}    & ACM MM2022      & Transformer       & -            & -           & 79.5          & 59.9         & 78.6          & 58.7         & 53.8    & -       & 42.0   \\
ViPT \cite{vipt}            & CVPR2023        & Transformer       & -            & -           & 83.5          & 61.7         & 83.5          & 61.7         & 65.1    & -       & 52.5   \\
MACFT \cite{MACFT}          & Sensors2023     & Transformer       & \color{blue}\textbf{80.1}         & \color{blue}\textbf{66.8}        & -             & -            & 85.7          & 62.2         & 65.3    & -       & 52.5   \\
RSFNet \cite{RSFNet}        & ISPL2023        & Transformer       & -            & -           & -             & -            & 86.3          & 62.2         & 65.9    & -       & 52.6   \\
DFMTNet \cite{DFMTNet}      & IEEE Sens.J23   & Transformer       & -            & -           & -             & -            & 86.2          & 63.6         & 65.1    & -       & 52.0   \\
TBSI \cite{tbsi}            & CVPR2023        & Transformer       & -            & -           & \color{blue}\textbf{85.3}    & \color{blue}\textbf{62.5}    & 87.1    & 63.7    & 69.2    & 65.7    & 55.6   \\
STTANet \cite{sttanet}      & TIM2024         & Transformer       & -            & -           & 82.5          & 60.2         & 85.5          & 63.2         & 66.7    & -       & 53.4   \\
STMT \cite{stmt}            & TCSVT2024       & Transformer       & -            & -           & 83.0          & 59.5         & 86.5          & 63.8         & 67.4    & 63.4    & 53.7   \\
TATrack \cite{tatrack}      & AAAI2024        & Transformer       & -            & -           & \color{blue}\textbf{85.3}          & 61.8         & 87.2          & 64.4         & 70.2    & \color{blue}\textbf{66.7}    & 56.1   \\
BAT \cite{bat}              & AAAI2024        & Transformer       & -            & -           & -             & -            & 86.8          & 64.1         & 70.2    & -       & 56.3   \\
TransAM \cite{amnet}        & TCSVT2024       & Transformer       & -            & -           & -             & -            & \color{blue}\textbf{87.7}    & \color{red}\textbf{65.5}         & 70.2    & 66.0    & 55.9   \\ \hline
CSTNet-small                & -               & Transformer       & \color{red}\textbf{82.2}   &\color{red}\textbf{69.3}          & 85.1         & 62.1          & 86.7         & 64.1         & \color{blue}\textbf{70.3}    & 66.3    & \color{blue}\textbf{56.4}   \\
CSTNet                      & -               & Transformer       & 77.5         & 66.0        & \color{red}\textbf{86.0} & {\color{red}\textbf63.5}         & \color{red}\textbf{88.4}          & \color{blue}\textbf{65.2}         & \color{red}\textbf{71.5 }   & \color{red}\textbf{67.9}    & \color{red}\textbf{57.2}   \\ \hline
\end{tabular}
\end{table*}

\subsection{Implementation Details}

\textcolor{black}{\textbf{Train: }}CSTNet is implemented on Ubuntu 20.04 using Python 3.9 and PyTorch 1.13.0. It is trained on a platform with two NVIDIA RTX A6000 GPUs over 20 epochs, which took roughly 6 hours. The LasHeR \cite{lasher} training set is used to train our model. In each epoch, we sample 60,000 image pairs, with a total batch size of 64. The total learning rate is set to $2\times10^{-5}$ for the backbone and $2e^{-6}$ for the feature fusion modules. It decays by a factor of 10 after 15 epochs. We use AdamW as the optimizer with a weight decay of $1e^{-4}$. The sizes of the template and search areas are set to 128x128 and 256x256, respectively. Overall, most of our training hyperparameters remain consistent with those of current trackers (such as \cite{vipt}, \cite{tbsi}, \cite{bat}) to ensure fairness in comparison. Our \textcolor{black}{JSCFM} and SFM modules are cascaded and inserted into the 4th, 7th, and 10th layers of the ViT. Here, our module insertion position is aligned with that of TBSI \cite{tbsi} to ensure fairness in comparison. CSTNet-small removes the \textcolor{black}{JSCFM} and SFM modules from the network and uses the CSTNet weights as pre-training weights, while its training settings remain the same as those of CSTNet.

\textcolor{black}{\textbf{Test: }LasHeR \cite{lasher} test set, RGBT210 \cite{rgbt210}, RGBT234 \cite{rgbt234}, VTUAV \cite{vtuav}, GTOT \cite{gtot}, and UniRTL \cite{unirtl} are used to evaluate our methods. }LasHeR \cite{lasher} \textcolor{black}{dataset} is categorized into 19 challenging attributes, including: No Occlusion (NO), Partial Occlusion (PO), Total Occlusion (TO), Hyaline Occlusion (HO), Out-of-View (OV), Low Illumination (LI), High Illumination (HI), Abrupt Illumination Variation (AIV), Low Resolution (LR), Deformation (DEF), Background Clutter (BC), Similar Appearance (SA), Thermal Crossover (TC), Motion Blur (MB), Camera Movement (CM), Frame Loss (FL), Fast Motion (FM), Scale Variation (SV), and Aspect Ratio Change (ARC). RGBT234 \cite{rgbt234} is a classic benchmark that consists of 234 sequences and features 12 challenging attributes: No Occlusion (NO), Partial Occlusion (PO), Heavy Occlusion (HO), Low Illumination (LI), Low Resolution (LR), Thermal Crossover (TC), Deformation (DEF), Fast Motion (FM), Scale Variation (SV), Motion Blur (MB), Camera Movement (CM), and Background Clutter (BC). RGBT210 \cite{rgbt210} is another classic RGB-T tracking benchmark containing 210 sequences. VTUAV \cite{vtuav} is a large-scale RGB-T tracking dataset from an aerial view, with its short-term tracking test subset consisting of 176 sequences. Lastly, GTOT \cite{gtot} is a classic RGB-T tracking benchmark that includes 50 short-term sequences.

\textcolor{black}{\textbf{Metrics}: }Following the metrics used by current most RGB-T trackers, precision rate (PR) and success rate (SR) are employed to evaluate different trackers on RGBT210 \cite{rgbt210}, RGBT234 \cite{rgbt234}, GTOT \cite{gtot}, \textcolor{black}{and UniRTL \cite{unirtl}}, and precision rate (PR), normalized precision rate (NPR) and success rate (SR) are employed to evaluate the trackers on LasHeR \cite{lasher}, and maximum precision rate (MPR) and maximum success rate (MSR) are employed on VTUAV \cite{vtuav}. 

\subsection{Comparison with State-of-the-art Methods}
\subsubsection{Comparison methods}

To evaluate the effectiveness of our method, we compare CSTNet and CSTNet-small with recent RGB-T trackers, including seven MDNet trackers: TFNet \cite{tfnet}, DMCNet \cite{dmcnet}, APFNet \cite{apfnet}, X-Net \cite{xnet}, AGMINet \cite{agminet}, DMSTM \cite{dmstm}, and CAT++ \cite{catpp}; three Siamese trackers: SiamCDA \cite{siamcda}, SiamMALL \cite{siammlaa}, and DFAT \cite{DFAT}; five DCF-based trackers: MFNet \cite{mfnet}, LSAR \cite{lsar}, CMD \cite{cmd}, MPT \cite{mpt}, and AMNet \cite{amnet}; and twelve transformer-based trackers: ProTrack \cite{protrack}, ViPT \cite{vipt}, MACFT \cite{MACFT}, RSFNet \cite{RSFNet}, DFMTNet \cite{DFMTNet}, TBSI \cite{tbsi}, STTANet \cite{sttanet}, STMT \cite{stmt}, BAT \cite{bat}, TATrack \cite{tatrack}, and TransAM \cite{amnet}.

\begin{figure}[t]
  \begin{minipage}[t]{0.5\linewidth}
    \centering
    \includegraphics[width=4.4cm]{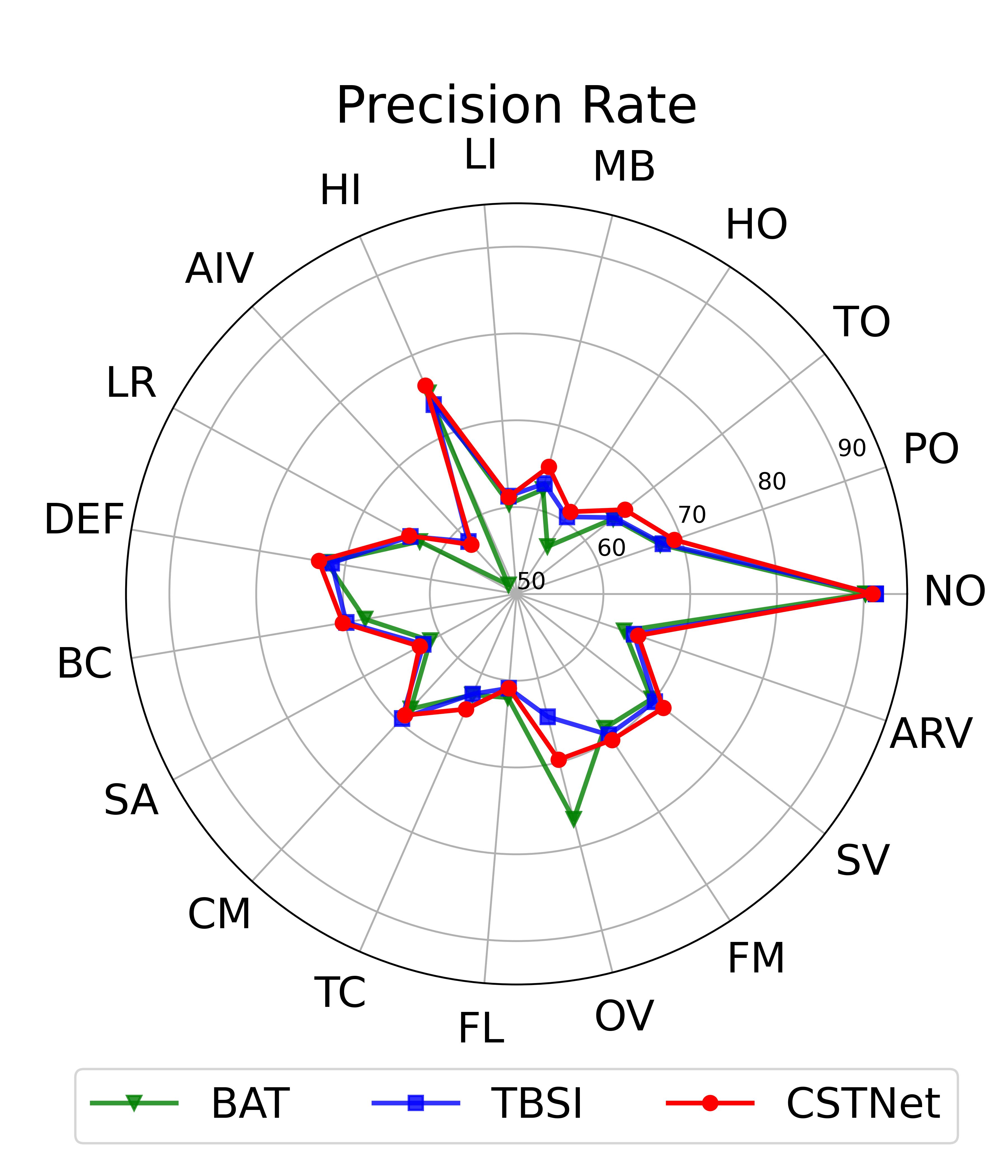}
  \end{minipage}%
  \begin{minipage}[t]{0.5\linewidth}
    \centering
    \includegraphics[width=4.4cm]{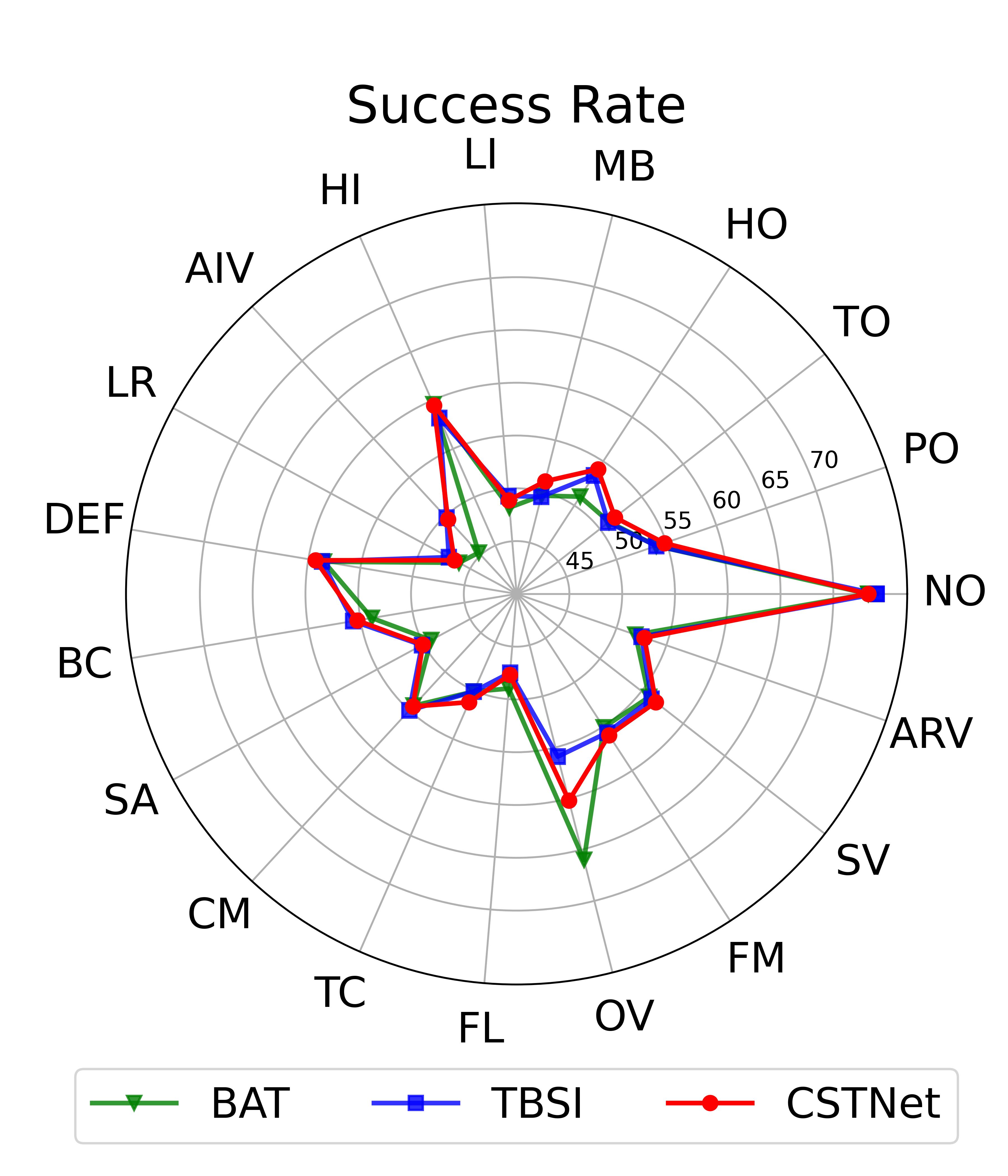}
  \end{minipage}
  \caption{The SR and PR scores of CSTNet, TBSI \cite{tbsi}, and BAT \cite{bat} under different attributes on LasHeR \cite{lasher}. }
  \label{fig:fig5}
\end{figure}

\begin{figure}[t]
  \begin{minipage}[t]{0.5\linewidth}
    \centering
    \includegraphics[width=4.4cm]{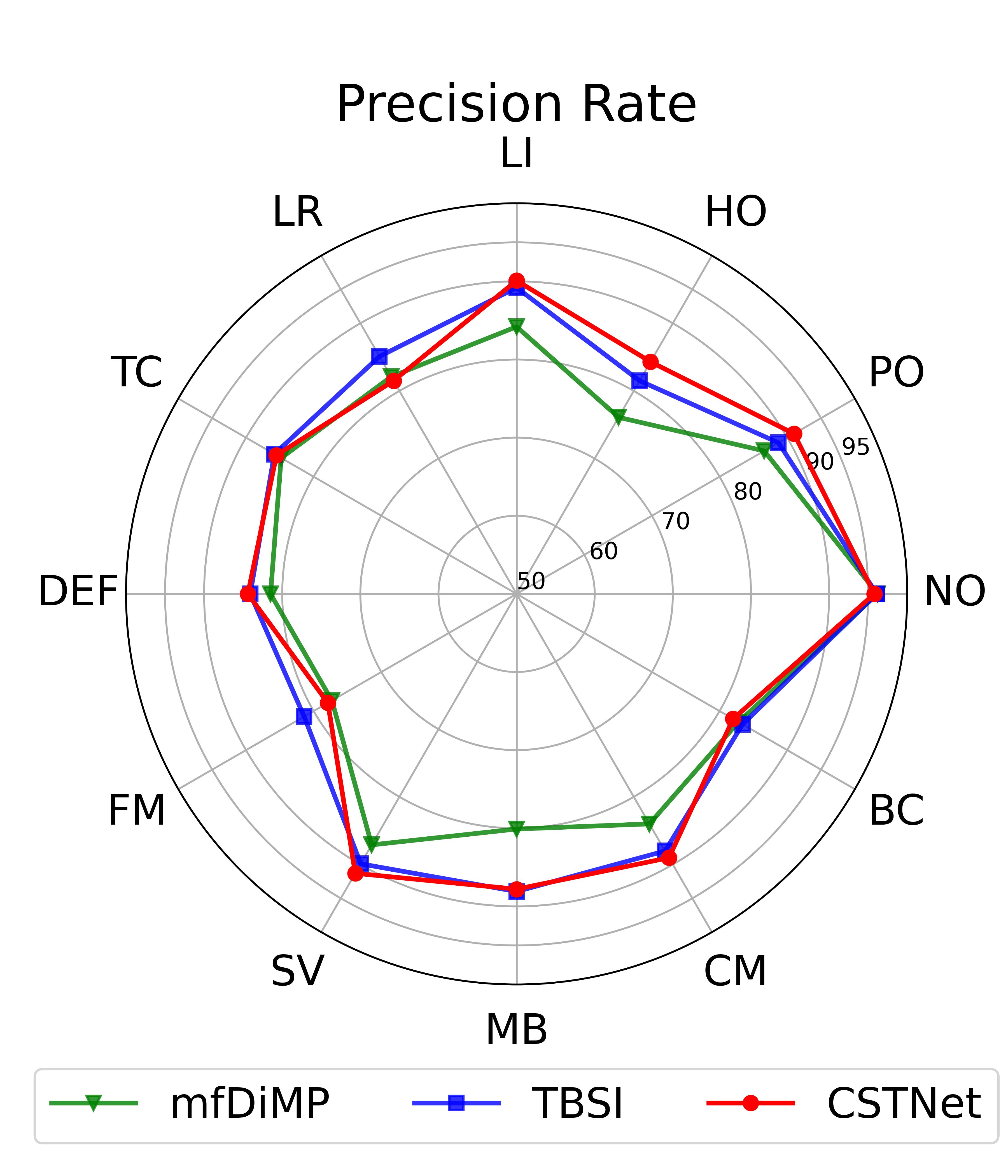}
  \end{minipage}%
  \begin{minipage}[t]{0.5\linewidth}
    \centering
    \includegraphics[width=4.4cm]{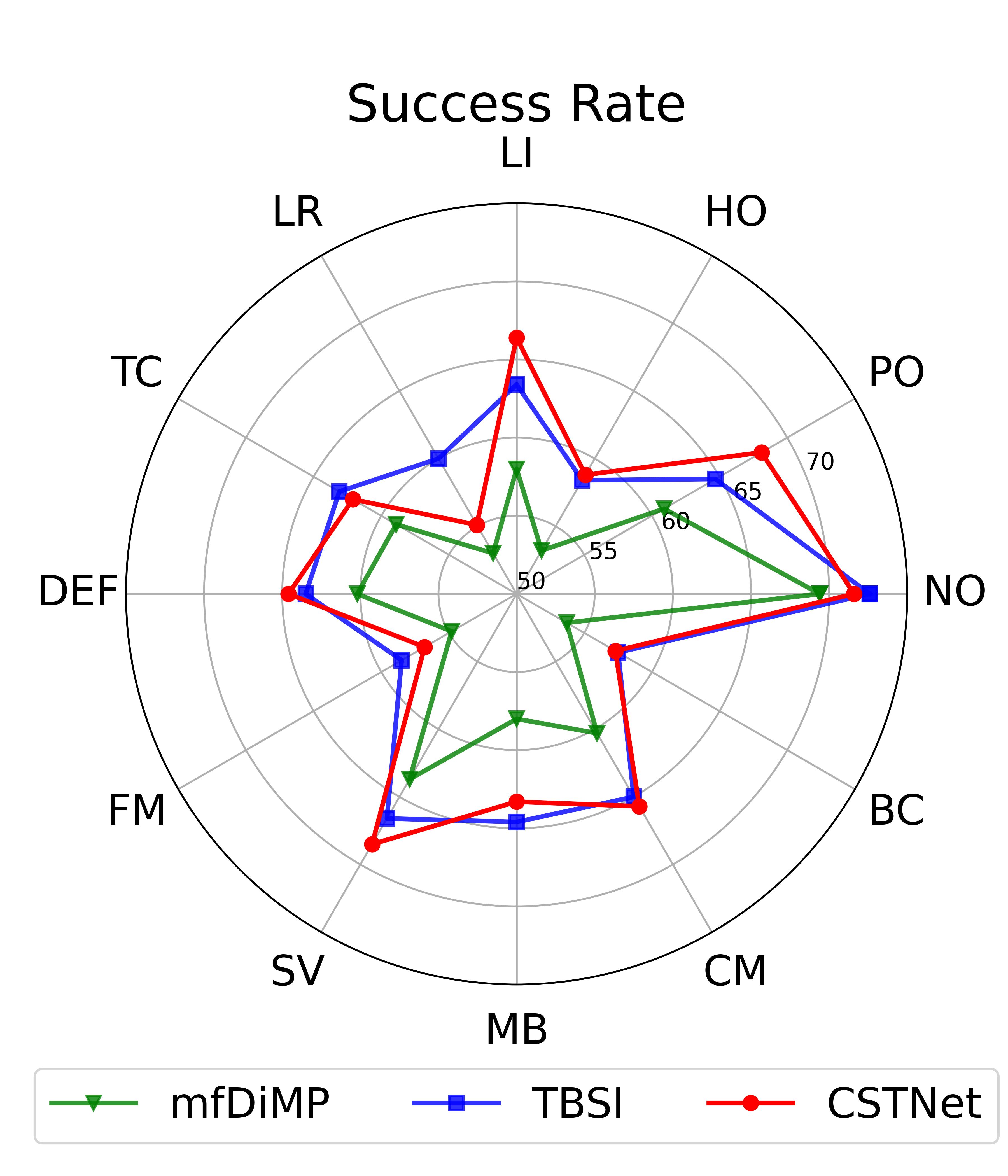}
  \end{minipage}
  \caption{The SR and PR scores of CSTNet, TBSI \cite{tbsi}, and mfDiMP \cite{mfdimp}, under different attributes on RGBT234 \cite{rgbt234}. }
  \label{fig:fig6}
\end{figure}

\begin{table}[t]
\caption{Comparison with state-of-the-art trackers on GTOT \cite{gtot}. The best two results
are shown in \color{red}{red} \color{black}{and} \color{blue}{blue} \color{black}{fonts}.}
\label{table result gtot}
\centering
\renewcommand{\arraystretch}{1.2}
\small %
\begin{tabular}{c|cc|cc}
\hline
       & \multirow{2}{*}{Baseline} & \multirow{2}{*}{Backbone} & \multicolumn{2}{c}{GTOT} \\ \cline{4-5} 
       &              &              & PR     & SR     \\ \hline
TFNet \cite{tfnet}   & MDNet           & VGG-M           & 72.4    & 88.6    \\
APFNet \cite{apfnet}  & MDNet           & VGG-M           & 73.7    & 90.5    \\
DMCNet \cite{dmcnet}  & MDNet           & VGG-M           & 73.8    & 90.9    \\
SiamCDA \cite{siamcda} & Siamese          & Res-50          & 87.7    & 73.2    \\
DFAT \cite{DFAT}    & Siamese          & Res-50          & 89.3    & 72.3    \\
SiamMLAA \cite{siammlaa}& Siamese          & Res-50          & 91.3    & 75.1    \\
LSAR \cite{lsar}    & DCF            & VGG-16          & 70.3    & 85.5    \\
CMD \cite{cmd}     & DCF            & Res-18          & 89.2    & 73.4    \\
AMNet \cite{amnet}   & DCF            & Res-50          & 91.1    & 76.1    \\
RSFNet \cite{RSFNet}  & Transformer        & ViT-B           & 92.1    & 75.3    \\
TransAM \cite{amnet}  & Transformer        & ViT-B           & \textbf{\color{blue}{92.9}}    & \textbf{\color{red}{77.4}}    \\ \hline
CSTNet-small      & Transformer        & ViT-B           & 91.5    & 75.4    \\
CSTNet         & Transformer        & ViT-B           & \textbf{\color{red}{93.0}}    & \textbf{\color{blue}{76.7}}    \\ \hline
\end{tabular}
\end{table}

\begin{table}[t]
\caption{Comparison with state-of-the-art trackers on UniRTL \cite{unirtl}. The best two results
are shown in \color{red}{red} \color{black}{and} \color{blue}{blue} \color{black}{fonts}.}
\label{table result unirtl}
\centering
\renewcommand{\arraystretch}{1.2}
\small %
\begin{tabular}{c|c|cc}
\hline

\multirow{2}{*}{} & \multirow{2}{*}{baseline} & \multicolumn{2}{c}{UniRTL} \\ \cline{3-4} 
                        &               & PR           & SR          \\ \hline
MANet \cite{manet}      & MDNet         & 37.8         & 45.3        \\
MFGNet \cite{mfgnet}    & MDNet         & 41.8         & 45.7        \\
ADRNet \cite{ADRNet}    & MDNet         & 45.1         & 48.8        \\
DFAT \cite{DFAT}        & MDNet         & 43.9         & 44.8        \\
APFNet \cite{apfnet}    & MDNet         & \color{blue}\textbf{45.4}         & 48.8        \\
TBSI \cite{tbsi}        & Transformer   & 43.2         & \color{blue}\textbf{51.3}        \\ \hline
CSTNet-small            & Transformer   & \color{blue}\textbf{45.4}         & 51.2        \\
CSTNet                  & Transformer   & \color{red}\textbf{46.1} & \color{red}\textbf{52.5}        \\ \hline
\end{tabular}
\end{table}

\begin{table}[t]
\caption{Comparison of parameters, Flops, and FPS on GPU RTX 3090Ti, GPU RTX 2080, and NVIDIA Jetson Xavier.}
\label{table params and flops}
\centering
\setlength{\tabcolsep}{2.2pt}
\renewcommand{\arraystretch}{1.2}
\small %
\begin{tabular}{c|cc|c|cc}
\hline
\multirow{2}{*}{} & \multicolumn{2}{c|}{3090Ti/2080}                       & Xavier & \multirow{2}{*}{Params}     & \multirow{2}{*}{Flops}     \\ \cline{2-4}
         & torch           & onnx      & onnx  &                 &                 \\ \hline
TBSI       & 60 / 28          & 71 / 40    & -   & 201.98M             & 82.5G              \\ \hline
CSTNet      & 62 / 36          & 74 / 44    & 21   & 144.00M             & 74.7G              \\
CSTNet-small   & \textbf{94} / \textbf{47} & \textbf{126} / \textbf{65} & \textbf{33}   & \textbf{92.12M} & \textbf{56.4G} \\ \hline
\end{tabular}
\end{table}

\begin{figure}[t]
  \begin{minipage}[t]{0.5\linewidth}
    \centering
    \includegraphics[width=4.8cm]{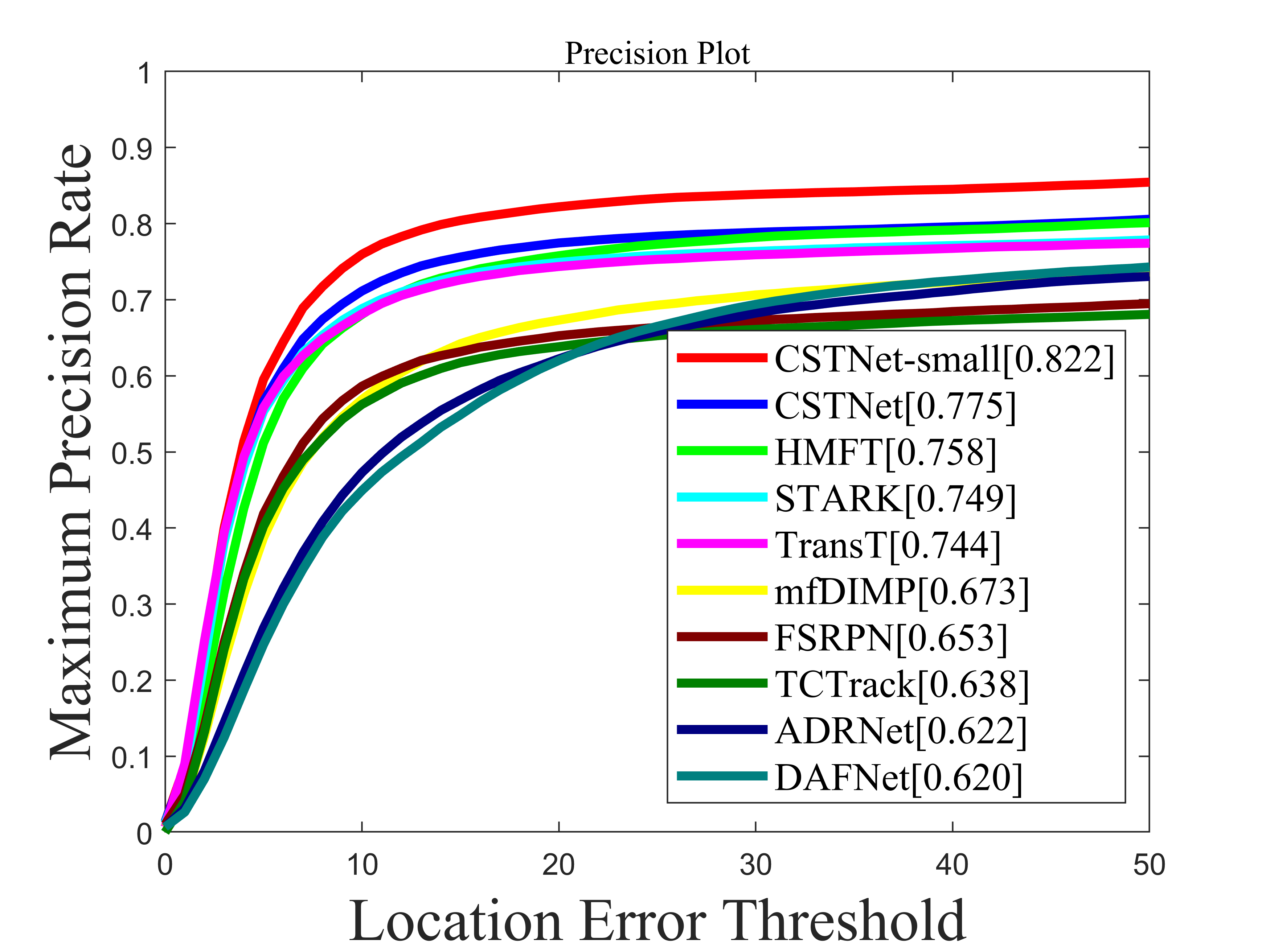}
  \end{minipage}%
  \begin{minipage}[t]{0.5\linewidth}
    \centering
    \includegraphics[width=4.8cm]{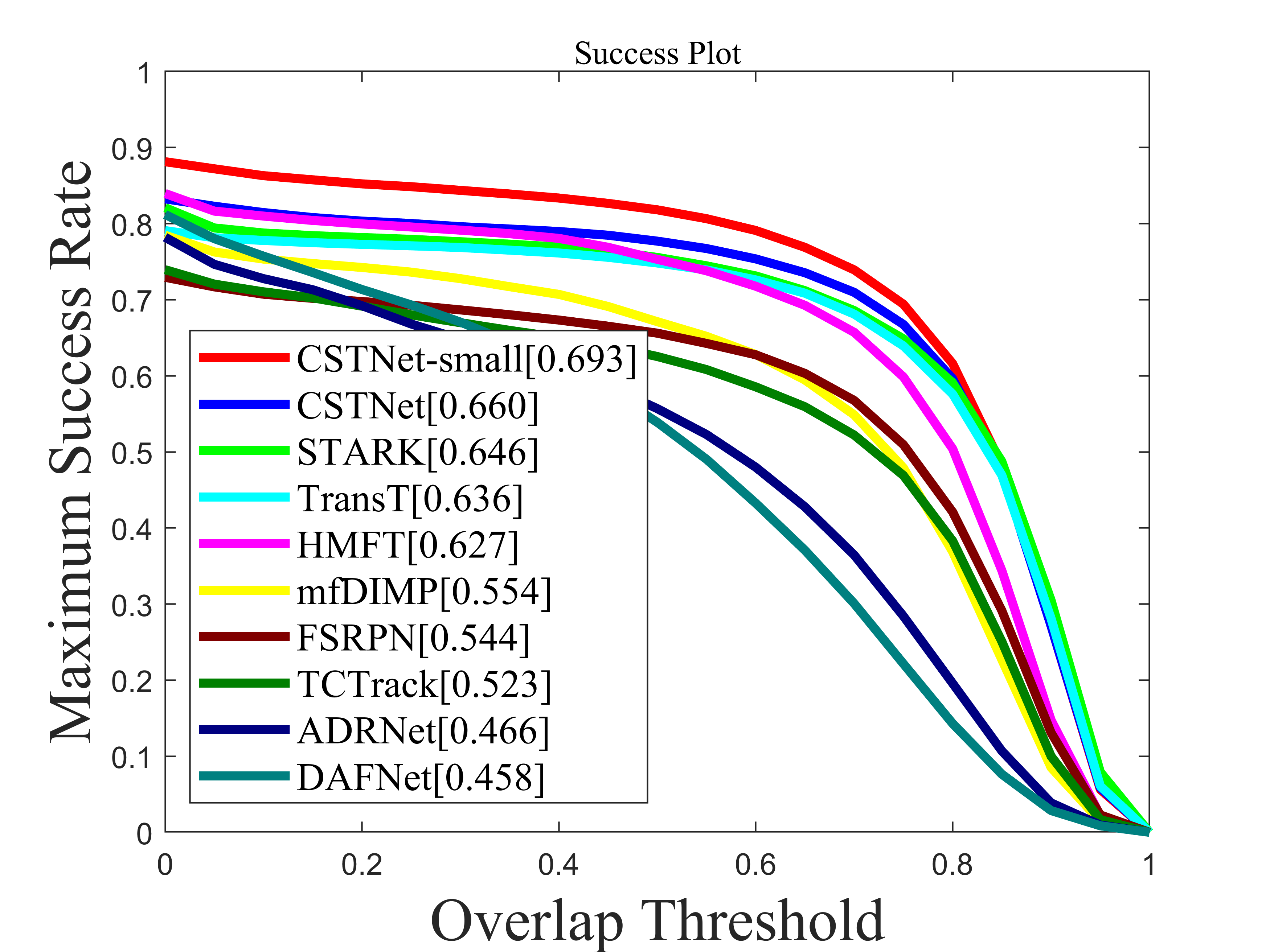}
  \end{minipage}
  \caption{Comparison with state-of-the-art trackers on VTUAV \cite{vtuav}. The trackers include HMFT \cite{vtuav}, Stark \cite{stark}, TransT \cite{transt}, mfDiMP \cite{mfdimp}, FSRPN \cite{vot19}, TCTrack \cite{tctrack}, ADRNet \cite{ADRNet}, \textcolor{black}{and} DAFNet \cite{dafnet}.}
  \label{fig:vtuav}
\end{figure}

\begin{table}[t]
\caption{Ablation studies of our proposed \textcolor{black}{JS}CFM and SFM modules in the tracking pipeline.}
\label{table ablation cfm and sfm}
\centering
\renewcommand{\arraystretch}{1.2}
\small %
\begin{tabular}{c|ccc}
\hline
Method     & PR      & NPR     & SR  \\ \hline
RGBT      & 70.0     & 66.3    & 56.0   \\ 
RGBT+\textcolor{black}{JS}CFM    & 70.5 (+0.5) & 67.0 (+0.7) & 56.5 (+0.5)\\
RGBT+SFM    & 68.5 (-1.5) & 64.8 (-1.5) & 55.1 (-0.9)\\
RGBT+\textcolor{black}{JS}CFM\&SFM  & \textbf{71.5 (+1.5)} & \textbf{67.9 (+1.6}) & \textbf{57.2 (+1.2)}\\ \hline
\end{tabular}
\end{table}

\begin{table}[t]
\caption{Ablation of the different methods for fusing search area cross-modal features.}
\label{table ablation addorcat}
\centering
\renewcommand{\arraystretch}{1.2}
\small %
\begin{tabular}{c|ccc}
\hline
Method & PR   & NPR  & SR    \\ \hline
concat & 70.0  & 66.4 & 55.1   \\
add   & \textbf{71.5}  & \textbf{67.9} & \textbf{57.2}   \\ \hline
\end{tabular}
\end{table}
 
\begin{table}[t]
\caption{Ablation of the different components of \textcolor{black}{JSCFM} module.}
\label{table ablation cfm}
\centering
\renewcommand{\arraystretch}{1.2}
\small %
\begin{tabular}{ccc|ccc}
\hline
\multicolumn{3}{c|}{\textcolor{black}{JSCFM}}                & \multirow{2}{*}{PR} & \multirow{2}{*}{NPR} & \multirow{2}{*}{SR} \\ \cline{1-3}
SE & \multicolumn{1}{c}{LSA} & \multicolumn{1}{c|}{GIM} &          &           &           \\ \hline
      &        &        & 69.5        & 65.8       & 55.8        \\
\checkmark &        &        & 70.5        & 66.8       & 56.3        \\
\checkmark & \checkmark  &        & 70.8        & 67.1       & 56.7        \\
\checkmark & \checkmark  & \checkmark  & \textbf{71.5}        & \textbf{67.9}       & \textbf{57.2}       \\ \hline
\end{tabular}
\end{table}

\begin{table}[t]
\caption{Ablation of the different components of SFM module.}
\label{table ablation sfm}
\centering
\renewcommand{\arraystretch}{1.2}
\small %
\begin{tabular}{ccc|ccc}
\hline
\multicolumn{3}{c|}{SFM}   & \multirow{2}{*}{PR} & \multirow{2}{*}{NPR} & \multirow{2}{*}{SR} \\ \cline{1-3}
LPU     & \multicolumn{1}{c}{CAM}  & \multicolumn{1}{c|}{CFN} &          &           &           \\ \hline
      &        &        & 70.5        & 67.0        & 56.5        \\
\checkmark &        &        & 70.6        & 67.1        & 56.5          \\
\checkmark & \checkmark  &        & 71.0        & 67.4        & 56.8        \\
\checkmark & \checkmark  & \checkmark  & \textbf{71.5}        & \textbf{67.9}        & \textbf{57.2}        \\ \hline
\end{tabular}
\end{table}

\begin{table}[t]
\caption{Ablation studies of different insert layers of our \textcolor{black}{JSCFM} and SFM modules.}
\label{table ablation layers}
\centering
\small %
\renewcommand{\arraystretch}{1.1}
\begin{tabular}{ccc|ccc}
\hline
\multicolumn{3}{c|}{Layers} & \multirow{2}{*}{PR} & \multirow{2}{*}{NPR} & \multirow{2}{*}{SR} \\ \cline{1-3}
4    & 7    & 10   &          &           &           \\ \hline
    &     &     & 70.0        & 66.3        & 56.0        \\
\checkmark    &     &     & 70.4        & 66.6        & 56.3        \\
\checkmark    & \checkmark    &     & 70.9        & 67.1        & 56.8        \\
\checkmark    & \checkmark    & \checkmark    & \textbf{71.5}        & \textbf{67.9}        & \textbf{57.2}        \\ \hline
\end{tabular}
\end{table}

\begin{table}[t]
\caption{Ablation studies of different pre-trained models.}
\label{table ablation pretrain}
\centering
\renewcommand{\arraystretch}{1.1}
\small %
\begin{tabular}{c|ccc}
\hline
Pre-train Model   & PR  & NP  & SR  \\ \hline
-          & 55.6 & 51.7 & 42.2 \\
ImageNet \cite{mae} & 59.1 & 56.0 & 47.3 \\
SOT \cite{ostrack} & 69.7 & 65.9 & 55.8 \\
RGBT \cite{tbsi}  & \textbf{71.5} & \textbf{67.9} &\textbf{57.2} \\ \hline
\end{tabular}
\end{table}
\subsubsection{Evaluation on LasHeR Dataset}

The trackers are evaluated on \textcolor{black}{the} LasHeR \cite{lasher}, and the results are reported in Table \ref{table rgbt234 and lasher reuslts}. CSTNet achieves state-of-the-art performance. Compared to other transformer-based trackers, CSTNet outperforms TATrack \cite{tatrack} and TransAM \cite{amnet} by 1.3\% in PR, 1.2\% and 1.9\% in NPR, and 1.1\% and 1.3\% in SR, respectively. Additionally, CSTNet's PR and SR are 1.3\% and 0.9\% higher than those of BAT \cite{bat}. Furthermore, CSTNet surpasses all other MDNet and Siamese methods. CSTNet-small achieves performance close to that of BAT. These results indicate that CSTNet provides better cross-modal feature interaction and overall performance.

In addition, we evaluate CSTNet, TBSI \cite{tbsi}, and BAT \cite{bat} on 19 attributes of the LasHeR \cite{lasher} dataset. As shown in Figure \ref{fig:fig5}, CSTNet outperforms other transformer-based trackers on many attributes. Compared to TBSI \cite{tbsi}, CSTNet demonstrates more competitive performance in challenges such as occlusion (HO, PO, and SO), thermal crossover (TC), out-of-view (OV), and motion blur (MB). When compared to BAT \cite{bat}, CSTNet excels in attributes like heavy occlusion (HO), abrupt illumination variation (AIV), and background clutter (BC). However, the PR and SR of CSTNet on attributes such as ARV, FL, SA, AIV, and LI are insufficient, indicating that CSTNet needs to further improve the model's robustness and feature discrimination.

\subsubsection{Evaluation on RGBT234 Dataset}
The trackers are evaluated on \textcolor{black}{the} RGBT234 \cite{rgbt234}, and the results are reported in Table \ref{table rgbt234 and lasher reuslts}. Compared to TransAM \cite{amnet}, CSTNet shows a 0.7\% improvement in PR. Furthermore, compared to BAT \cite{bat}, CSTNet achieves a significant improvement of 1.6\% in PR and 1.1\% in SR. Additionally, CSTNet outperforms CSTNet-small by 1.7\% in PR and 1.1\% in SR. These results demonstrate that CSTNet establishes a new state-of-the-art benchmark.

We further evaluate our proposed CSTNet, TBSI \cite{tbsi}, and mfDiMP \cite{mfdimp} on 12 attributes of the RGBT234 \cite{rgbt234} dataset. As shown in Figure \ref{fig:fig6}, CSTNet outperforms other trackers on most attributes. However, it exhibits insufficient success rates in heavy occlusion (HO), low resolution (LR), background clutter (BC), and fast motion (FM). This suggests that CSTNet faces challenges in addressing issues such as missing key features and interference from background elements.

\subsubsection{Evaluation on RGBT210 Dataset}

The trackers are evaluated on \textcolor{black}{the} RGBT210 \cite{rgbt210}, and the results are reported in Table \ref{table rgbt234 and lasher reuslts}. CSTNet surpasses all other RGB-T trackers in performance. Specifically, the PR and SR of CSTNet are 0.7\% and 1.0\% higher than TBSI \cite{tbsi}, respectively. Furthermore, CSTNet achieves an SR of 63.5\%, surpassing STMT \cite{stmt} and TATrack \cite{tatrack} by 4.0\% and 1.7\%, respectively. These experimental findings demonstrate that CSTNet outperforms all other trackers and establishes state-of-the-art performance.

\subsubsection{Evaluation on VTUAV-ST Dataset}

The trackers were evaluated on the VTUAV dataset \cite{vtuav}, with the results presented in Figure \ref{fig:vtuav}. Both CSTNet and CSTNet-small outperform all other RGB-T trackers. Specifically, CSTNet demonstrates improvements of 1.7\% in \textcolor{black}{MPR} and 3.3\% in \textcolor{black}{MSR} compared to HMFT \cite{vtuav}. Furthermore, CSTNet-small shows enhancements of 4.7\% in MPR and 3.3\% in MSR relative to CSTNet. These experimental results indicate that CSTNet-small surpasses all other trackers, achieving state-of-the-art performance.

\subsubsection{Evaluation on GTOT Dataset}

The trackers are evaluated on \textcolor{black}{the} GTOT \cite{gtot}, with results reported in Table \ref{table result gtot}. CSTNet outperforms all other RGB-T trackers in \textcolor{black}{PR}. Compared to RSFNet \cite{RSFNet}, CSTNet improves PR and \textcolor{black}{SR} by 0.9\% and 1.4\%, respectively. The experimental results demonstrate that both CSTNet and CSTNet-small achieve state-of-the-art performance.

\subsubsection{Evaluation on UniRTL Dataset}
The trackers are evaluated on the UniRTL \cite{gtot}, with results reported in Table \ref{table result unirtl}. CSTNet outperforms all other RGB-T trackers in both PR and SR. Compared to TBSI \cite{tbsi}, CSTNet improves PR and SR by 2.9\% and 1.2\%, respectively. Compared to APFNet \cite{apfnet}, CSTNet improves PR and SR by 0.7\% and 3.7\%, respectively. The experimental results demonstrate that both CSTNet and CSTNet-small achieve state-of-the-art performance.

\subsection{Parameters and FLOPs}

We also evaluate the parameters, \textcolor{black}{FLOPs}, and FPS of the models. The results are shown in Table \ref{table params and flops}. Compared to our baseline \cite{tbsi}, our CSTNet and CSTNet-small achieve better performance, reducing approximately 25\% and 50\% of the parameters and 10\% and 30\% of the \textcolor{black}{FLOPs}, respectively. Our CSTNet and CSTNet-small achieve 62 \textit{fps} and 94 \textit{fps}, respectively, on the GPU RTX 3090Ti. In addition, we deploy our CSTNet and CSTNet-small on NVIDIA Jetson Xavier. \textcolor{black}{Both models} run at speeds of 21 \textit{fps} and 33 \textit{fps}, demonstrating the feasibility of our method in practical applications.

\textcolor{black}{However, compared to TBSI \cite{tbsi}, the speedup of CSTNet was not significantly improved, despite the fact that it demonstrated fewer parameters and FLOPs. We believe this may be due to CSTNet's architecture using a large number of residual connections, which may result in a reduction in speed.}

\subsection{Ablation Studies}
To investigate the individual influence of different components, we conduct ablation studies on CSTNet and report the results for PR, NPR\textcolor{black}{,} and SR on LasHeR \cite{lasher}.

\subsubsection{Component Analysis}

First, we evaluate the effect of summing the RGB and TIR features and the contributions of the \textcolor{black}{JSCFM} and SFM modules on the tracker’s performance. RGBT denotes the addition of the RGB and TIR features from the search area output by the shared ViT backbone, followed by predicting the target state. The baseline pretrained model is referenced in \cite{tbsi}. As shown in Table \ref{table ablation cfm and sfm}, the contributions of the \textcolor{black}{JSCFM} module and the SFM module result in improvements of 0.5\% and 1.0\% in terms of PR, 0.7\% and 0.9\% in terms of NPR, and 0.5\% and 0.7\% in terms of SR, respectively. Moreover, they achieve total improvements of 1.5\%, 1.6\%, and 1.2\% in terms of PR, NPR, and SR, respectively. Additionally, the 1-2\% performance degradation of CSTNet when using only the SFM module confirms the discussion in TBSI \cite{tbsi} that cross-attention may not be suitable for RGB-T tracking. In fact, the cross-modal features integrated through the \textcolor{black}{JSCFM} module can yield positive gains when combined with the cross-attention-based SFM module. This further underscores the importance of cascading the \textcolor{black}{JSCFM} and SFM modules, illustrating their interdependence rather than merely improving features separately. This distinction sets our method apart from other cross-attention-based approaches. Overall, both our proposed \textcolor{black}{JSCFM} and SFM modules are essential.

Additionally, we investigate the impact of concatenating versus adding the RGB and TIR features of the search area output from the backbone on tracking performance. The method referred to as concat is utilized by TBSI \cite{tbsi}, where the RGB and TIR features of the search area are concatenated, followed by a $1\times1$ convolution for dimensionality reduction. As shown in Table \ref{table ablation addorcat}, adding the RGB and TIR features in the search area yields better performance. Our approach avoids the feature representation bottleneck associated with dimensionality reduction through $1\times1$ convolution, enabling the model to benefit from the direct addition of RGB and TIR features in the search area.

\subsubsection{Ablation of \textcolor{black}{JSCFM} module}

We evaluate the contributions of each component of the \textcolor{black}{JSCFM}. As shown in Table \ref{table ablation cfm}, when the \textcolor{black}{JSCFM} module is removed from the model, the PR and SR are 69.5\% and 55.8\%, respectively. Adding the SE module improves the PR by 1.0\% and the SR by 0.5\%. Then, using the LSA module improves performance by 0.3\% for SR and 0.4\% for PR. Next, using the GIM module to integrate global features results in an improvement of 0.7\% for PR and 0.5\% for SR. Overall, all three modules in \textcolor{black}{JSCFM} module contribute to improving the performance of our model.

\subsubsection{Ablation of SFM module}
The contribution of each component to the SFM is measured. Table \ref{table ablation sfm} shows the results.
The contribution of the LPU module \textcolor{black}{results in} a 0.1\% improvement in PR. The CAM module improves the PR by 0.4\% and SR by 0.3\%. The CFN module improves the PR by 0.5\% and \textcolor{black}{the} SR by 0.4\%. Overall, all three components of the SFM module are effective.

\subsubsection{Inserting Layers of \textcolor{black}{JSCFM} and SFM Modules}
In addition, we conduct ablation experiments on different \textcolor{black}{insertion} layers of our proposed \textcolor{black}{JSCFM} and SFM modules. The results are reported in Table \ref{table ablation layers}. Inserting the \textcolor{black}{JSCFM} and SFM modules in the 4th layer of the ViT improves the PR by 0.4\% and the SR by 0.3\%. Inserting the modules in the 7th and 10th layers of \textcolor{black}{the} ViT further improves the performance of the tracker. Overall, inserting our \textcolor{black}{JSCFM} and SFM modules into these three layers effectively achieves feature fusion at different stages and improves model performance.

\subsubsection{Ablation of pre-trained models}
\label{section: pretrain models}
We explore the impact of different pre-trained models on our proposed CSTNet in terms of performance. \textcolor{black}{Both} CSTNet without pretraining and CSTNet with ImageNet pretraining are unable to converge in 20 epochs; therefore we increase their learning rate by 10 times. The results are shown in Table \ref{table ablation pretrain}, CSTNet with RGB-T pretraining achieves the best performance. It outperforms CSTNet with SOT pretraining by 1.8\% for PR and 1.4\% for SR. The RGB-T pre-training weight\textcolor{black}{s} achieves better parameter initialization of our model, reducing our training epochs to one-third compared to ViPT \cite{vipt}, TATrack \cite{tatrack}, and STMT \cite{stmt} and BAT \cite{bat}. 
This shows that using the RGB-T pre-training weight can further improve the performance of the tracker.

The use of RGB-T pretraining weight\textcolor{black}{s} is not intended to gain an inappropriate performance advantage; rather, we hope to promote the development of RGB-T trackers through a better \textcolor{black}{pretraining} weight\textcolor{black}{s}, as in the case of \cite{ostrack}.

\subsection{Visualization}

\subsubsection{Heatmap}

\begin{figure}[t]
	\centering
    \includegraphics[width=8.8cm]{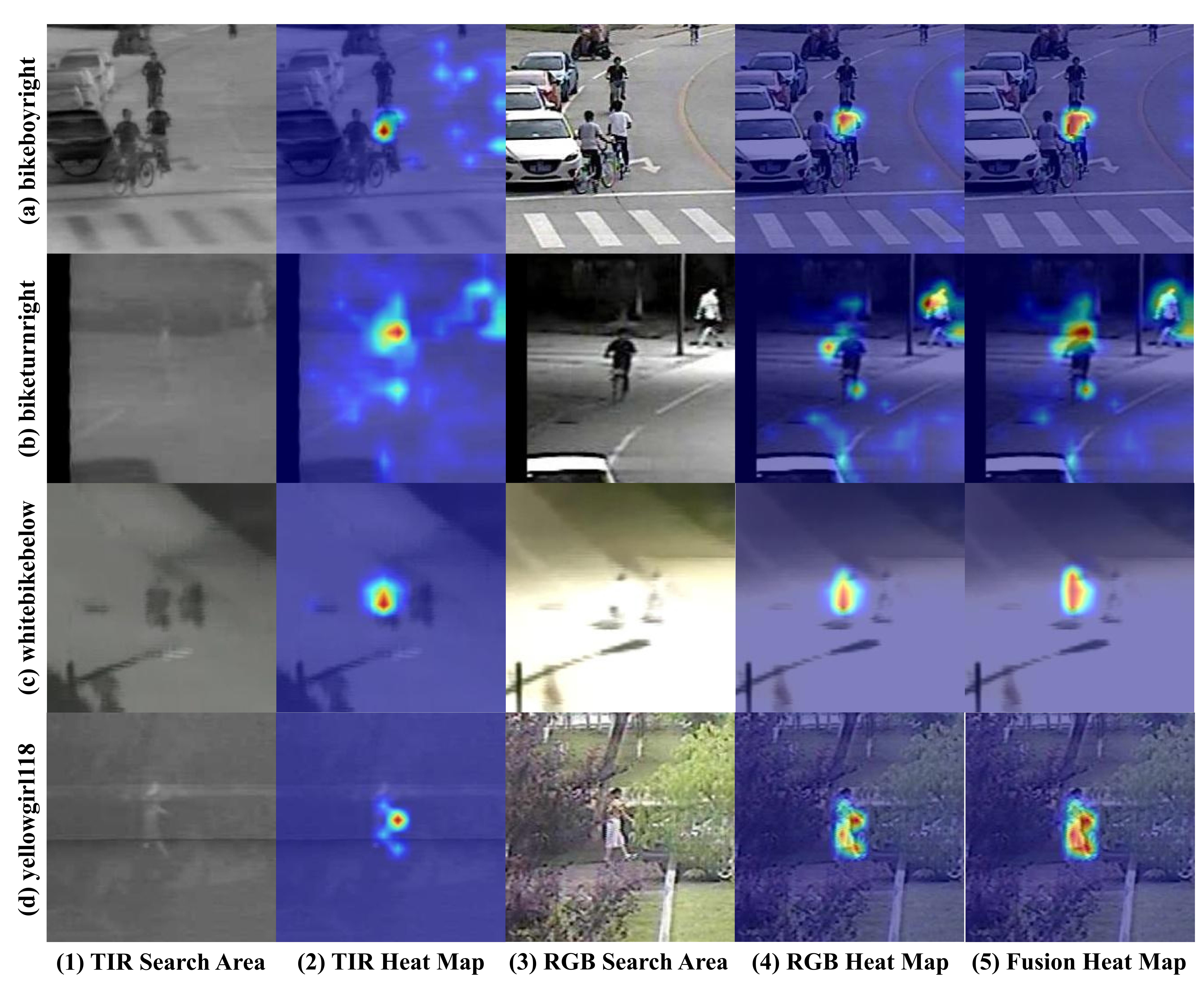}
	\caption{Visualization of heat maps of RGB feature, TIR feature and fusion feature of the search area on LasHeR \cite{lasher} based on Grad-CAM. The \textit{Fusion Heat Map} represents the heatmap of the sum feature of the RGB and TIR modalities. The fusion feature's heat map is only displayed in RGB modality. (a) bikeboyright (frame 301), (b) biketurnright (frame 46), (c) whitebikebelow (frame 165), (c) yellowgirl118 (frame 65).}
	\label{fig:fig7}
\end{figure}

\begin{figure}[t]
	\centering
    \includegraphics[width=8.8cm]{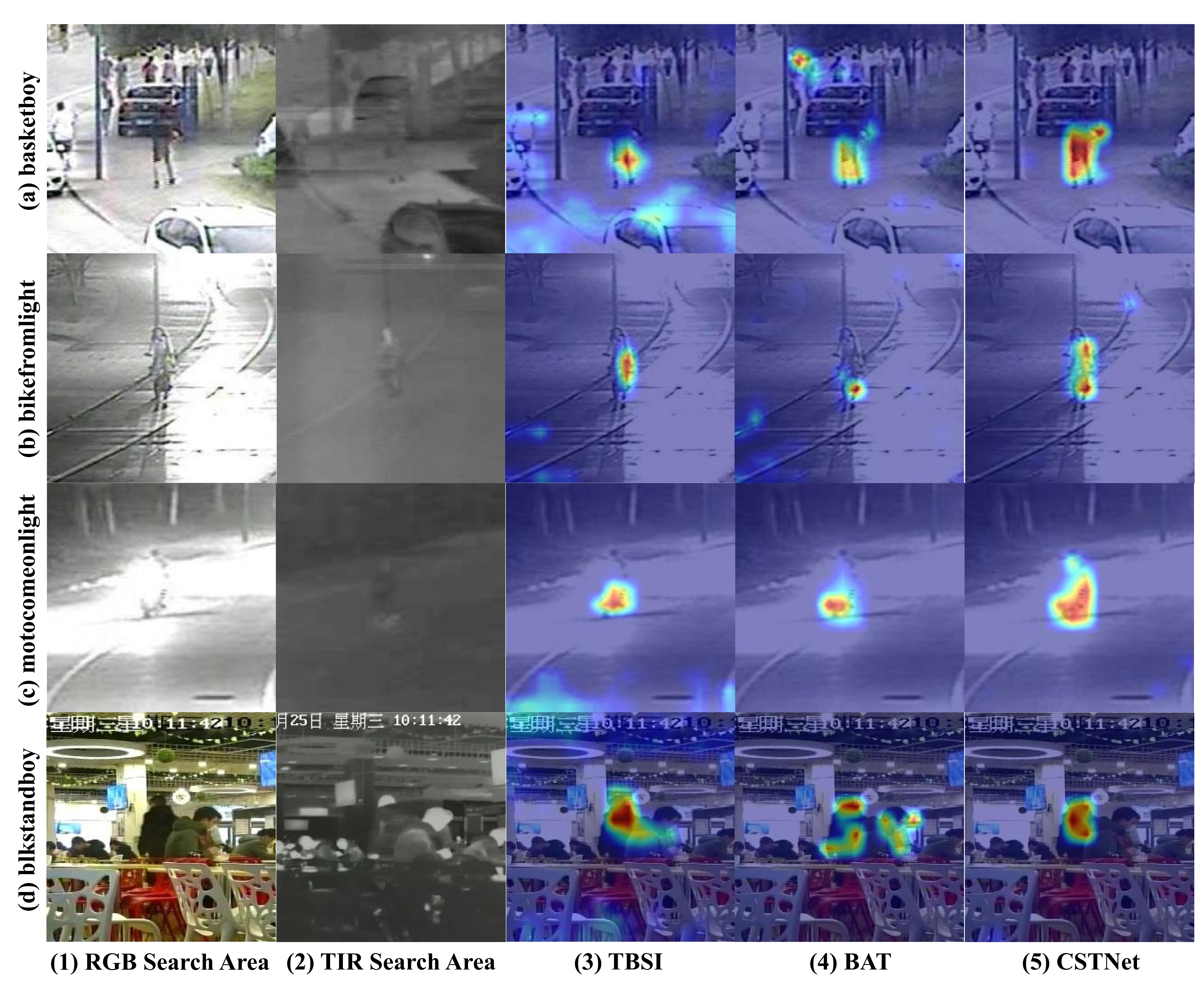}
	\caption{Visualization of heat maps of the fusion feature of CSTNet, TBSI \cite{tbsi}, and BAT \cite{bat} on LasHeR \cite{lasher} based on Grad-CAM. The fusion feature's heat map is only displayed in RGB modality. (a) basketboy (frame 48), (b) bikefromlight (frame 138), (c) motocomeonlight (frame 31), (c) blkstandboy (frame 128).}
	\label{fig:fig8}
\end{figure}

We explore how CSTNet leverages RGB and TIR modalities to enhance model performance. As illustrated in Figure \ref{fig:fig7}, during the bikeboyright sequence, a typical daytime traffic scenario, CSTNet effectively integrates RGB and TIR modalities to improve target state prediction. In the biketurnright sequence, a nighttime scenario, although the RGB features are disturbed, the model successfully predicts the target state by relying on the fused features guided by the TIR modality. Furthermore, in the whitebikebelow and yellowgirl118 sequences, CSTNet enhances target state prediction under high illumination and partial occlusion by complementing RGB and TIR features. This demonstrates CSTNet's capability for effective cross-modal feature interaction.

Additionally, we compare the cross-modal feature interaction performance of CSTNet with other Transformer-based methods, namely TBSI \cite{tbsi} and BAT \cite{bat}, across four typical sequences from the LasHeR \cite{lasher} dataset, as shown in Figure \ref{fig:fig8}. CSTNet achieves direct interaction between the original semantics of RGB and TIR through \textcolor{black}{the} channel and spatial fusion of multimodal features. In the basketboy, bikefromlight, and motocomeonlight sequences, CSTNet shows richer utilization of cross-modal appearance features. In the blkstandboy sequence, it exhibits superior discrimination for similar appearances. Overall, CSTNet highlights the advantages of direct cross-modal feature interaction in enhancing model performance.

\subsubsection{Tracking Results}

\begin{figure*}[t]
	\centering
    \includegraphics[width=18.cm]{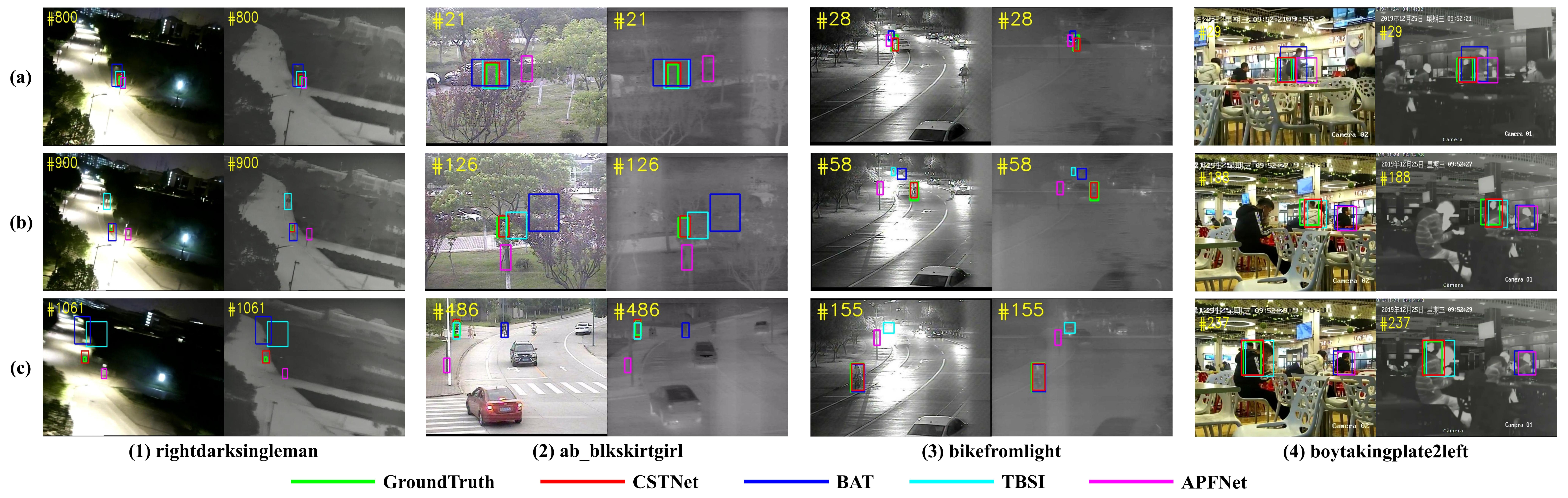}
	\caption{Visualized comparisons of CSTNet with TBSI \cite{tbsi}, BAT \cite{bat}, and APFNet \cite{apfnet} on four sequences from LasHeR \cite{lasher} dataset. (1) rightdarksingleman, (2) ab\_blkskirtgirl, (3) bikefromlight, (4) boytakingplate2left. (a) result example 1. (b) result Example 2. (c) result Example 3.}
	\label{fig:fig9}
\end{figure*}

We select four representative sequences from the LaSHeR \cite{lasher} dataset to show some of the results of CSTNet and other previous state-of-the-art RGB-T trackers. As shown in Figure \ref{fig:fig9}, in the rightdarksingleman sequence, when other trackers briefly lose the target due to low light and camera shake, our proposed CSTNet still tracks the target stably, demonstrating the \textcolor{black}{superior} robustness of CSTNet. In the ab\_blkskirtgirl sequence, when the target is frequently occluded, other trackers usually lose the target briefly, while our CSTNet does not. In the bikefromlight sequence, when the scene is highly illuminated, our CSTNet accurately distinguishes the target from the surrounding clutter, resulting in better performance. In the boytakingplate2left sequence, our CSTNet better integrates the RGB and TIR features of the target, \textcolor{black}{leading to improved} performance in partial occlusion and similar appearance challenges. Overall, our proposed CSTNet demonstrates stronger discriminative ability and better tracking performance.

We provide additional visualization and analysis in the supplementary material.

\subsubsection{Failure Cases}
We present some typical failure cases of CSTNet and CSTNet-small on four representative sequences in the LaSHeR \cite{lasher} benchmark, as shown in Figure \ref{fig:failure cases}. In the shinycarcoming sequence, when strong illumination and occlusion occur, our methods experience tracking drift. In the whiteridingbike sequence, the complete occlusion poses a challenge to our methods. The whitesuvturn and leftchair sequences demonstrate the limitations of our methods in dealing with objects of similar appearance under different illumination conditions. Overall, our method needs further improvement to adapt to challenges such as strong illumination, occlusion, and similar objects.

\begin{figure*}[t]
	\centering
    \includegraphics[width=18.cm]{figure_failure_cases.jpg}
	\caption{Visualizion of failure cases of CSTNet and CSTNet-small with TBSI \cite{tbsi}, and BAT \cite{bat} on four sequences from LasHeR \cite{lasher} dataset. (1) shinycarcoming, (2) whiteridingbike, (3) whitesuvturn, (4) leftchair. (a) failure case 1. (b) failure case 2. (c) failure case 3.}
	\label{fig:failure cases}
\end{figure*}

\color{black}

\section{Discussion}

\subsection{Discussion of CSTNet-small}

We argue that the efficiency of CSTNet-small comes from being retrained based on the CSTNet pre-trained model. Specifically, CSTNet models the common semantics in the multimodal features of a target through the \textcolor{black}{JSCFM} and SFM modules, which are in its backbone. This feature modeling capability is contained in the backbone due to the \textcolor{black}{large} number of parameters and efficient structure. \textcolor{black}{T}he proposed \textcolor{black}{JSCFM} and SFM modules may play a role in constraining the multimodal semantic modeling. Therefore, when CSTNet-small is retrained, the CSTNet-small model initialized with CSTNet weights inherits part of the modeling capability \textcolor{black}{from} the backbone. Although the absence of these modules restricts CSTNet-small's semantic interactions at different levels, resulting in lower performance than CSTNet, its performance \textcolor{black}{remains} competitive. We provide additional discussion and analysis in the supplementary material.

\subsection{Practical Applicability}

The GTOT \cite{gtot}, RGBT210 \cite{rgbt210}, and RGBT234 \cite{rgbt234} datasets reflect urban scenes in the real world. The VTUAV \cite{vtuav} dataset reflects urban or outdoor scenes from the perspective of UAVs. Our methods achieve state-of-the-art performance on these dataset benchmarks, demonstrating their applicability in these scenarios. In addition, Table \ref{table params and flops} shows that both CSTNet and CSTNet-small achieve real-time speeds on the NVIDIA Jetson Xavier embedded device, proving their deployability in uncrewed vehicles. Overall, our proposed method is suitable for current potential RGB-T tracking applications. Moreover, our method can be deployed not only in fixed visible light and infrared monitoring \textcolor{black}{equipment}, but also in monitoring equipment on uncrewed vehicles. \textcolor{black}{Specifically, our method provides the coordinates of the target in the image, and then obtains its coordinates relative to the vehicle. In addition, in some search and rescue missions, the Global Positioning System (GPS) coordinates of the target can be obtained based on the motion status of the vehicle, the GPS information of the vehicle, and the motion status of the target, thereby achieving more accurate positioning.}

\subsection{Limitation}
In this work, although our \textcolor{black}{JSCFM} and SFM modules effectively achieve direct interaction between RGB and TIR modality features, there are still some limitations. For CSTNet, our method focuses on learning multimodal discriminative appearance features of the target \textcolor{black}{while ignoring the spatio-temporal features of the target in different modalities \cite{isj_st1}\cite{isj_st2}}. This limits the adaptability of CSTNet when the target appearance changes significantly. \textcolor{black}{Our method lacks deep mining of difficult samples \cite{isj_gmmt}. Furthermore, our method does not consider further integration with prompt \textcolor{black}{layers} or adapter \textcolor{black}{layers} \cite{isj_prompt1}}. 
For CSTNet-small, although we effectively reduce model parameters and FLOPs, and \textcolor{black}{increase} speed, \textcolor{black}{it is uncertain} whether this process can be refined through knowledge distillation. Similarly, it remains unclear whether this paradigm of removing modules and retraining can be extended to the optimization of models with stringent requirements on speed, parameters, and FLOPs.

\subsection{Future Work}
Firstly, we will explore the modeling of multimodal spatio-temporal features within the tracker to enhance the model's adaptability to changes in target appearance. Secondly, we will \textcolor{black}{investigate} the paradigm of model retraining combined with knowledge distillation to determine whether model performance can be improved through the pre-training of auxiliary modules.

\section{Conclusion}
In this paper, we present a new RGB-T tracker called CSTNet. CSTNet leverages channel and spatial feature fusion to facilitate direct interaction between RGB and TIR features. \textcolor{black}{It} includes two novel cross-modal feature interaction modules: \textcolor{black}{Joint Spatial and Channel Fusion Module (JSCFM)} and Spatial Feature Module (SFM). The \textcolor{black}{JSCFM} module enhances the joint channel representation of RGB and TIR features, also modeling features at multiple levels and integrating both fused and original features globally. The SFM module focuses on spatial mutual modeling between RGB and TIR features, enabling joint spatial and channel integration of multimodal features. Comprehensive experiments demonstrate that our proposed CSTNet achieves state-of-the-art results on challenging RGB-T tracking benchmarks \textcolor{black}{and the effectiveness of direct feature interaction}. Additionally, we introduce a smaller version of the tracker, CSTNet-small, which maintains high performance in RGB-T tracking by omitting the \textcolor{black}{JSCFM} and SFM modules, utilizing the CSTNet weights for pre-training. We evaluate the parameters, \textcolor{black}{FLOPs}, and speed of both CSTNet and CSTNet-small, successfully deploying them on an embedded computing device. Our experiments confirm that both models meet the real-time requirements for practical applications.

\section{Acknowledgments}
This research is funded by the National Natural Science Foundation of China, grant number 52371350 and by the National Key Research and Development Program of China, grant number 2023YFC2809104.

\bibliography{references}
\bibliographystyle{IEEEtran}

\section{Nomenclature Table}
The abbreviations used in this article are in alphabetical order are as shown in Table \ref{table nomenclature}.

\begin{table*}[]
\caption{\textcolor{black}{The abbreviations used in this article are in alphabetical order as follows:}}
\label{table nomenclature}
\centering
\renewcommand{\arraystretch}{1.1}
\begin{tabular}{c|l|c|l}
\hline
Nomenclature & Referred to                               & Nomenclature & Referred to                          \\ \hline
AIV          & Abrupt Illumination Variation             & MSR          & Maximum Success Rate                 \\
ARC          & Aspect Ratio Change                       & NO           & No Occlusion                         \\
BC           & Background Clutter                        & NPR          & Normalized Precision Rate            \\
CAM          & Cross-Attention Module                    & OV           & Out-Of-View                          \\
CFN          & Convolutional Feedforward Network         & PO           & Partial Occlusion                    \\
CM           & Camera Movement                           & PR           & Precision Rate                       \\
CSTNet       & Channel And  Spatial  Transformer Network & SA           & Similar Appearance                   \\
DEF          & Deformation                               & SE           & Squeeze-And-Excitation Module        \\
FL           & Frame Loss                                & SFM          & Spatial Fusion Module                \\
FM           & Fast Motion                               & SR           & Success Rate                         \\
GIM          & Global Integration Module                 & SV           & Scale Variation                      \\
HI           & High Illumination                         & TC           & Thermal Crossover                    \\
HO           & Hyaline Occlusion                         & TIR          & Thermal Infrared                     \\
JSCFM        & Joint Spatial And Channel Fusion Module   & TO           & Total Occlusion                      \\
LI           & Low Illumination                          & UAV          & Uncrewed Aerial Vehicle              \\
LPU          & Local Perception Unit                     & USV          & Uncrewed Surface Vehicle             \\
LR           & Low Resolution                            & UV           & Underwater Vehicle                   \\
LSA          & Local Spatial Aggregation                 & ViT          & Vision Transformer                    \\
MB           & Motion Blur                               & FLOP         & Floating Point Operations Per Second \\
MHSA         & Multi-Head Self-Attention                 & GPS          & Global Positioning System            \\
MPR          & Maximum Precision Rate                    & FPS          & Frames Per Second                    \\ \hline
\end{tabular}
\end{table*}

\end{document}